\documentclass{article} 
\usepackage{preprint,times}

\usepackage{amsmath,amsfonts,bm}

\def\eqref#1{equation~\ref{#1}}

\def\1{\bm{1}}

\DeclareMathAlphabet{\mathsfit}{\encodingdefault}{\sfdefault}{m}{sl}
\SetMathAlphabet{\mathsfit}{bold}{\encodingdefault}{\sfdefault}{bx}{n}

\usepackage{hyperref}
\usepackage{url}
\usepackage{booktabs}       
\usepackage{amsfonts}       
\usepackage{nicefrac}       
\usepackage{microtype}      
\usepackage{xcolor}         
\usepackage{amsmath, bm}
\usepackage{wrapfig}
\usepackage{graphicx}
\usepackage{colortbl}
\usepackage{amssymb}
\usepackage[noabbrev,nameinlink,capitalize]{cleveref}
\usepackage{subcaption}
\usepackage{multirow} 

\usepackage{algorithm}
\usepackage{algorithmic}

\definecolor{Red}{rgb}{0.768, 0.054, 0.054}
\definecolor{Blue}{rgb}{0.43, 0.65, 0.86}
\definecolor{Green}{rgb}{0,0.4,0.7}
\definecolor{hotpink}{rgb}{1.0, 0.41, 0.71}
\definecolor{brown}{rgb}{0.59, 0.29, 0.0}
\definecolor{darkpastelgreen}{rgb}{0.01, 0.75, 0.24}
\definecolor{celestialblue}{rgb}{0.29, 0.59, 0.82}
\definecolor{ceruleanblue}{rgb}{0.16, 0.32, 0.75}
\definecolor{goldenrod}{rgb}{0.85, 0.65, 0.13}
\definecolor{navyblue}{rgb}{0.0, 0.0, 0.5}
\definecolor{coolgrey}{rgb}{0.55, 0.57, 0.67}
\definecolor{darkseagreen}{rgb}{0.56, 0.74, 0.56}
\definecolor{darkturquoise}{rgb}{0.0, 0.81, 0.82}
\definecolor{berryred}{rgb}{0.79, 0.25, 0.14}
\definecolor{teagreen}{rgb}{0.81,0.94,0.75}
\definecolor{lightgrey}{rgb}{0.5,0.5,0.5}
\definecolor{purple}{rgb}{0.35,0.25,0.55}
\hypersetup{
    colorlinks=true,
    citecolor=teal,
    linkcolor=Red,
    urlcolor=Green,
    }

\title{A Night at the LLM Zoo: Data Conditioned Flow Matching for Generating Pretrained LLM Weights}
\author{Scott Biggs \\
Khoury College of Computer Science\\
Northeastern University, Silicon Valley\\
\texttt{biggs.s@northeastern.edu}\\
\And
Bedionita Soro \\
MLAI Lab\\
KAIST\\
\texttt{sorobedio@kaist.ac.kr}\\
\And
Saumya Gupta \\
Khoury College of Computer Science\\
Northeastern University, Boston\\
\texttt{gupta.saumy@northeastern.edu}\\
\AND
Bruno Andreis \\
Slater Lab\\
University of Oxford\\
\texttt{bruno.andreis@eng.ox.ac.uk}\\
\And
Sung Ju Hwang\thanks{Corresponding author.} \\
MLAI Lab, KAIST \& DeepAuto.ai\\
\texttt{sungju.hwang@kaist.ac.kr}\\
}

\begin{document}

\maketitle
\begin{abstract}
A challenge to Weight Space Learning as it expands from task specific and vision models to Large Language Models (LLMs) is the absence of a well curated and analyzed LLM population. To address this, we train such a model zoo for application to Weight Space Generation by systematically varying the data mixture composition between web text, books, mathematics, coding, and French across populations of $51$M, $124$M, and $355$M parameter GPT-2 Models, each tier with count $N=100$. We analyze these populations by cosine similarity, functional diversity, Leave-One-Mixture-Out modeling, and variance decomposition. We show that the zoo contains diverse models whose variation is structured by training mixture composition. We embed the training mixture composition to condition a Flow Matching model of \textit{complete} pretrained LLM weights in Gram-PCA compressed weight space. By demonstrating zero-shot generalization to held out mixtures, measuring Continuous Ranked Probability Score skill, as well as finding minimal changes in downstream test-set perplexity and general capability benchmark scores, we show that our generator generalizes on the simplex of data mixtures to produce models that perform similarly to those in the zoo \textit{ex nihilo}. Finally, we produce all source code and artifacts to support the continued expansion of Weight Space Learning and Generation into the LLM domain. 
\end{abstract}
\section{Introduction}

The success of modern deep neural networks relies heavily on large-scale computation, and the past decade has produced a rapidly growing population of pretrained models, from computer vision systems to large language models (LLMs). Despite the substantial computational cost invested in training these models, the information contained in their learned parameters remains comparatively underexplored for understanding how neural networks evolve during training or for constructing new models. A central difficulty is that publicly available checkpoints are highly heterogeneous: they often differ in architecture, scale, initialization, training data, and optimization procedure, while many of these details are only partially documented. As a result, collections of existing models do not provide a controlled population from which the structure and dynamics of LLM weight space can be readily studied.

A recent surge of research on weight-space learning addresses this problem by treating the parameters of trained neural networks as data~\citep{han2026survey, wang2026position}. However, existing studies remain concentrated on relatively small vision and classification models, or on parameter-efficient representations such as Low-Rank Adaptation (LoRA) weights, leaving the full parameter space of pretrained language models comparatively unexplored ~\citep{doctolora, huang2026loragen}. This limitation can be attributed to the  difficulty of learning meaningful structure from collections of LLMs whose training conditions and architectural differences are uncontrolled. Among the few works that extend full-weight generation to natural-language models, DeepWeightFlow~\citep{dwflow} trains a population of 100 BERT  encoders from scratch and models their parameters; however, these models are task-specific regressors trained on a single supervised objective rather than pretrained language models which consider data from multiple diverse sources. Similarly, \citet{soro2026lsmerge} studies full LLM weights in the context of model merging and highlights the difficulty of obtaining compact and meaningful representations of large language-model parameters. Whether systematic and exploitable structure exists in the full parameter space of models trained under a general language-modeling objective therefore remains an open question.

We argue that addressing this question requires a controlled model zoo specifically designed for studying pretrained LLM weight space. Rather than assembling unrelated checkpoints from public repositories, such a zoo should systematically control the factors that generate variation across models, making it possible to associate changes in model parameters with known changes in pretraining conditions. Motivated by this challenge, we construct a structured zoo of pretrained language models containing more than $300$ models trained under systematically varying pretraining data compositions. This controlled population provides a foundation for studying how LLM weights vary across training distributions, characterizing the geometry of their parameter space, and ultimately determining whether this structure is sufficiently regular to support the analysis and generation of full language-model parameters. 

Furthermore, we train conditional flow matching models~\citep{lipman2023flow, liu2023flow}
to generate the weights of GPT style models~\citep{radford2019language} at 51M, 124M and
355M parameters, conditioned on the pretraining data mixture. Earlier work on parameter
generation has mostly focused on small vision models or low rank
adapters~\citep{Peebles2022, soro2025d2nwg, wang2025scaling, charakorn2025, huang2026loragen}, whereas we
generate complete pretrained language models. Following~\citet{dwflow}, we compress each
zoo with Gram PCA, train the flow in this compressed space, and decode the samples back
into full GPT-2 weights, which we evaluate directly without any further training. The generated models perform close to the original trained models and generalize well to data mixtures held out from training. They stay closer to the real models than a nearest neighbor retrieval baseline, and their CRPS
skill~\citep{Gneiting2007StrictlyPS} shows that conditioning on the mixture clearly helps. Our results show that the generated weight spaces generalize zero shot to new data mixtures.

The primary contributions of our work are:
\begin{itemize}
    \item We build a controlled zoo of 300 GPT-2 models at 51M, 124M and 355M parameters, in which the only factor that varies between models is the composition of the pretraining data.
    \item We show that the weights in this zoo are structured by data mixture. Models trained on the same mixture are strongly aligned, models trained on different mixtures are nearly orthogonal, and the mixture explains more than 85\% of the variance among repeatedly drawn models.
    \item We train a conditional flow matching generator that produces complete pretrained language models for unseen data mixtures across model parameter scales. The generated models achieve an average perplexity change of 4.1 to 5.1\%, compared to 5.8 to 6.0\% for a nearest neighbor retrieval baseline, and perform on par with our real models on HellaSwag~\citep{zellers2019hellaswagmachinereallyfinish}.
\end{itemize}

\section{Related Work}

\paragraph{Weight-space learning.} A growing body of work treats trained parameters as a data modality in their own
right \citep{han2026survey, wang2026position}. Early studies predicted model properties such as accuracy directly from weights \citep{unterthiner2020,eilertsen2020}. Later work designed architectures that respect the symmetries of
weight space, including permutation-equivariant layers \citep{navon2023,zhou2023}, graph metanetworks~\citep{kofinas2024, lim2024}, and set-based encoders \citep{andreis2023}. A parallel line learns weight representations with autoencoders, yielding hyper-representations used for both property prediction
and sampling \citep{schurholt2021, skde, sane}. These methods are developed and evaluated on small vision classifiers and implicit neural representations, not on pretrained language models. In addition, weight-space learning relies on populations of models rather than isolated checkpoints. Controlled model zoos vary seeds and hyperparameters while holding architecture fixed \citep{ModelZoosDataset2022}, and later work studies how zoo
composition affects learned representations \citep{falk2025zoo} or extends the setting to heterogeneous public checkpoints \citep{falk2025hub}. These zoos are primarily vision-based. We adopt the same controlled setting for language models, varying only the pretraining data composition rather than optimization or other hyperparameters.

\paragraph{Weight generation.}
Generating parameters instead of optimizing them dates back to hypernetworks~\citep{ha2017} and to graph hypernetworks that predict parameters for unseen architectures \citep{knyazev2021}. Generative models over checkpoints followed, including diffusion over optimization trajectories \citep{Peebles2022}, diffusion~\citep{soro2025d2nwg, wang2025scaling} weights, autoregressive parameter generation \citep{soro2025igpg}, and flow matching in a re-basined weight space \citep{dwflow}. In the language domain, generation is largely confined to low-rank adapters, where a hypernetwork maps a task description or a batch of unlabeled prompts to LoRA matrices~\citep{charakorn2025, draganddrop, doctolora, huang2026loragen}. Such methods produce a small update on top of a frozen backbone and therefore do not model pretrained parameters themselves. The exception is \citet{dwflow}, which trains a population of BERT-sized encoders and generates their full weights, but that population is trained on a single supervised regression task rather than on general-purpose language modeling.

\paragraph{Controlled suites and data composition.} Open suites release language models and intermediate checkpoints to study training: MultiBERTs varies random seed \citep{sellam2022}, Pythia varies scale while keeping data and ordering fixed \citep{biderman2023}, and PolyPythias extends this to multiple pretraining runs \citep{polypythias}; LLM360~\citep{liu2023llm360} and OLMo~\citep{olmo20242olmo2furious} similarly release open training artifacts. These suites are designed to isolate factors such as scale, seed, or training step and therefore largely keep the pretraining distribution fixed. The complementary literature instead varies data composition: DoReMi learns domain weights \citep{xie2023}, data mixing laws relate mixture proportions to language-modeling loss \citep{ye2024}, RegMix predicts performance from sampled mixtures \citep{liu2025regmix}, and DCLM compares curation strategies under a standardized recipe \citep{li2024dclm}. These works study how data composition affects performance, but not how it is reflected in the resulting parameters. We retain the weights and ask whether the pretraining mixture is recoverable from weight space.

\section{Methods}
\label{sec:methods}
\subsection{Problem Setup \& Notation}
\label{sec:setup}
Let $\mathcal{T}_1, \ldots, \mathcal{T}_K$ denote $K$ pretraining corpora over distinct
text domains (see Table~\ref{tab:domains} for details), where $K = 5$. Let $\pi \in \Delta^{K-1} = \Delta^4$ be a mixture over these
domains, with $\pi_d$ denoting the weight on domain $\mathcal{T}_d$. Each pretraining mixture $\pi$ produces a distribution of model parameters $\theta$, where all models have the same number of parameters $D$. Given a model zoo $\Theta = {\theta_i}{i=1}^N$, where each model $\theta_i$ was pretrained on a different mixture $\pi_i$, we study how the model weights change as the pretraining mixture changes. Our goal is to determine whether these changes follow a regular pattern that can be utilized to learn a conditional generator $f\psi(z \mid \pi)$ to generate plausible model parameters for mixtures that were not seen during training. In practice, rather than generating parameters directly in the high-dimensional parameter space $\mathbb{R}^D$, we learn this mapping in the latent space introduced in Section~\ref{sec:zoo_encoding}.

\subsection{LLM Zoo Construction}
\label{sec:zoo}

Each zoo holds $N=100$ members over $85$ distinct mixtures. Twenty are \emph{anchors}:
the five one-hot vertices of $\Delta^{4}$, each replicated over four independent branches
at the same $\pi$. The remaining eighty are \emph{singletons}, at unique interior mixtures
drawn from $\mathrm{Dirichlet}(1)$ and rejection-sampled at a minimum pairwise $L_1$ gap
of $0.15$. The anchors are the only mixtures drawn more than once, hence the only place a
within-mixture quantity can be estimated.

We construct the zoo to capture variation in the weight spaces directly attributable to pretraining compositions such that all members share architecture,
optimizer, schedule, token budget and initialization, and only $\pi_i$ and a branch seed
vary. For the first $(1-\beta)S$ of $S$ optimizer steps a single model trains on a shared
stream, forming a \emph{trunk}, which is then copied into $N$ branches; branch $i$ trains
for the remaining $\beta S$ steps on $\pi_i$, under a schedule spanning
the full trajectory so that warmup does not restart. We set $\beta=0.15$ and a budget of
$20$ tokens per parameter \citep{3600270.3602446}, and build three zoos at $51$M (Mini),
$124$M (Small) and $355$M (Medium) parameters, the latter two being published GPT-2
configurations ~\citet{radford2019language}. 
Across these scales only a single mixture plan is drawn, so across scales member $i$ carries an identical $\pi_i$. Additional details on Zoo construction can be found throughout Appendix~\ref{app:designing_zoo}. 


\subsection{Zoo Encoding and Weight-Space Representation}
\label{sec:zoo_encoding}

We follow the Gram-PCA compression technique for extending weight space generation beyond $\mathcal{O}(100\text{M})$ developed by ~\cite{dwflow}, representing a given zoo tier with population $N=100$ in dimension $k=N-1=99$ codes for zoo diversity analysis, and $k=50$ codes for generative modeling experiments. Details are explored in Appendix~\ref{app:compression_algo}~\citep{dwflow}. 


We choose rank $k=N-1=99$ for reporting population data, while generative experiments are done at $k=50$. The cost of this truncation is measured at $\sim2\%\Delta \text{PPL}$ in Table~\ref{tab:gen-ladder}. Due to our method of zoo training, detailed in Section~\ref{sec:zoo} and Appendix~\ref{app:designing_zoo}, we do not need to re-basin the population (see Appendix~\ref{app:beta_lmc})\citep{pmlr-v119-frankle20a}. Finally, a notable limitation of this Gram-PCA technique is that it must be fitted on every member, even in LOMO or zero-shot settings. Appendices~\ref{app:k_choice}, \ref{app:beta_lmc} and~\ref{app:transductive}
expand on these points. 

\subsection{LLM Zoo Weight Generation}
\label{sec:llm_zoo_gen}
We model $p(z\mid\pi)$ with a rectified flow \citep{liu2023flow}, the
linear-interpolant case of conditional flow matching \citep{lipman2023flow}. For a member
code $z_1$ and Gaussian source $z_0\sim\mathcal{N}(0,\sigma_0^2 I_k)$ we take
$z_t=(1-t)z_0+tz_1$ and regress a velocity field on the constant target $z_1-z_0$,
\begin{equation}
  \mathcal{L}(\psi)=\mathbb{E}_{t\sim\mathcal{U}[0,1],\,(z_1,\pi)\sim\mathcal{Z},\,z_0}
  \big\|v_{\psi}\big(z_t,t,e_{\psi}(\pi)\big)-(z_1-z_0)\big\|_2^{2},
\label{eq:cfm}
\end{equation}
where $v_{\psi}$ is an MLP and $e_{\psi}$ a two-layer network embedding $\pi$ into a
vector concatenated to the velocity network's input \citep{3504035.3504518}, trained jointly
with the flow \citep{dwflow, soro2025d2nwg, saragih2025flowlearnflowmatching}; a learned null token with conditioning
dropout gives classifier-free guidance \citep{ho2021classifierfree}. Sampling integrates
$\dot z_t=v_\psi$ by the explicit Euler method, and the resulting latent PCA-code is decoded and evaluated
without further optimizer steps. Further details and ablations can be found in Appendix~\ref{app:cfm_details}.

\subsection{LLM Zoo and Generated LLM Diversity}
\label{sec:diversity}

\subsubsection{LLM Zoo Diversity}
\label{sec:llm_zoo_diversity}

We adopt the reporting conventions of vision model zoos  developed by \cite{ModelZoosDataset2022, unterthiner2020, eilertsen2020}. However, some are adjusted due to our trunk-branch approach. Our zoo also contains only a single optimizer and model architecture, only varying the data supply, $\pi \in \Delta^4$, across the population at a given scale. 

\paragraph{Branch deltas.}
Every $\theta_i$ inherits the trunk, so $\|\theta_i-\theta_j\|$ are small relative to 
$\theta_i$. We work instead with the \emph{branch delta}
$d_i=\theta_i-\theta_{\text{trunk}}$, member $i$'s task vector
\citep{ilharco2023editing}, and read all pairwise quantities off the $N\times N$ Gram
$G_{ij}=\langle d_i,d_j\rangle$ through 
\begin{equation}
\|\theta_i - \theta_j\|^2 = G_{ii} + G_{jj} - 2G_{ij},
\qquad
\|\theta_i - \theta_{\text{trunk}}\|^2 = G_{ii},
\label{eq:pop-gram}
\end{equation}
in which the reference cancels. Displacements, not raw weights, are a numerical
requirement here, not a convenience: on raw weights the first identity is catastrophic
cancellation.

\paragraph{Cosine pools.}
The pairwise diversity statistic is $\cos_{ij}=G_{ij}/\sqrt{G_{ii}G_{jj}}$, read against
the null for independent directions in $\mathbb{R}^D$, $|\cos|\sim D^{-1/2}$. We report it in four disjoint pools, created from anchor pairs sharing a $\pi$, anchor pairs with different
$\pi$, interior (or singleton) to interior, and anchor to singleton. These pairings' displacement scales differ materially enough that they cannot be simply combined. Raw cosines confound structure induced by $\pi$ with the drift inherited from
the trunk, which we refer to as the mean displacement $\bar d=N^{-1}\sum_i d_i$. Projecting out this shared drift yields the centered Gram,
\begin{equation}
\tilde G \;=\; P G P, \qquad P = I - \tfrac{1}{N}\mathbf{1}\mathbf{1}^{\!\top},
\label{eq:pop-centred}
\end{equation}
we report both forms when possible.

\paragraph{Between- and within-$\pi$ variance.}

The four replicate branches per anchor give a genuine within-group term, admitting a one-way decomposition with $d=5$ groups of $s=4$,
\begin{equation}
\mathrm{SS}_B = s\sum_{g}\|m_g - \bar m\|^2,
\qquad
\mathrm{SS}_W = \sum_{g}\sum_{r}\|\theta_{gr} - m_g\|^2,
\qquad
\mathrm{SS}_T = \mathrm{SS}_B + \mathrm{SS}_W,
\label{eq:pop-anova}
\end{equation}
for $m_g$ the group mean and $\bar m$ the population mean, making
$\eta^2=\mathrm{SS}_B/\mathrm{SS}_T$ the share of anchor variation attributable to mixture identity \citep{fisher1925statistical}. Since $\mathrm{SS}_B$ carries $d-1$ degrees of
freedom of within-group noise, $\eta^2$ is biased upward. Treating the five vertices as fixed effects, we report the unbiased estimate of
the between-mixture variance and the corresponding variance fraction
\citep{searle1992variance},,
\begin{equation}
\hat\sigma^2_{\text{dom}}
  = \frac{\mathrm{SS}_B - (d-1)\,\mathrm{MS}_W}{ds},
\qquad
\mathrm{MS}_W = \frac{\mathrm{SS}_W}{d(s-1)},
\qquad
\eta^2_{\text{corr}} = \frac{\hat\sigma^2_{\text{dom}}}
                            {\hat\sigma^2_{\text{dom}} + \mathrm{MS}_W},
\label{eq:pop-anova-corr}
\end{equation}
alongside $\eta^2$ to address this. A Gram is a sum over coordinates, so \eqref{eq:pop-anova} is exactly additive over disjoint coordinate regions; we quote the transformer blocks separately from the embeddings throughout, since the embedding share of $D$ falls by $\sim\!3.4\times$ across our three scales.

\paragraph{Leave-One-Mixture-Out.}~\eqref{eq:pop-anova} needs repeated draws at a single $\pi$, so it is defined only
at the five anchors, the vertices of $\Delta^4$. These are the most mutually distinct mixtures
the design contains. The interior of $\Delta^4$ admits the out-of-sample form of the same question:
withhold a mixture, predict its branch delta from $\pi$ alone, and score that prediction
against simply predicting the mean of the retained members
\citep{Allen1974TheRB,Arlot_2010}. We call this Leave-One-Mixture-Out (LOMO),
\begin{equation}
R^2_{\mathrm{LOMO}} \;=\; 1 \;-\;
\frac{\sum_{F} \bigl\| d_F - \hat{d}_F^{\,(-F)} \bigr\|^2}
     {\sum_{F} \bigl\| d_F - \bar{d}^{\,(-F)} \bigr\|^2},
\label{eq:pop-loo}
\end{equation}
where $\hat{d}^{\,(-F)}$ is an ordinary least-squares fit on the rows outside fold $F$ and
$\bar{d}^{\,(-F)}$ is the mean of those same rows, both refit for every fold.

The folds are mixtures, not members. An anchor's four branches share a $\pi$ exactly, so
withholding one member leaves three siblings at an identical key and the prediction
degenerates to retrieval --- the leakage grouped cross-validation exists to prevent whenever
records are clustered. 


We fit several methods of varied complexity to this objective and compare their performances against our flow generator: a linear fit on $\pi$, a linear fit on $\pi$ with its entropy feature $H(\pi)$, and a quadratic fit on $\pi$. Through classical
multidimensional scaling from $\mathbb{R}^k$, these same statistics transfer to $\mathbb{R}^D$ \citep{Gower1966SomeDP} (see Appendix~\ref{app:more-lomo}).

\paragraph{Population spectra.}
We report two distinct spectra: the five \emph{centered domain means}, whose rank is
bounded by $d-1=4$ by construction, and all $N$ \emph{members} at $k=99$. Both are
summarized by the participation-ratio effective rank: $\mathrm{eff}(\lambda) \;=\; \frac{\left(\sum_a \lambda_a\right)^2}{\sum_a \lambda_a^2}$, 
quoted absolutely with its $k$ stated. We do not report the normalized variant eff/$k$, because it is not $k$-invariant. 

\paragraph{Functional diversity.}

Neither weight nor latent space metrics directly capture functional diversity ~\citep{wang2024neural, zeng2025generative}. Therefore, we also report diversity on downstream tasks in the form of next-token-prediction disagreement rates under greedy decoding ~\citep{germann-2003-greedy}.

\subsubsection{Generated LLM Diversity}
\label{sec:gen_diversity}

Throughout, a \emph{cell} is one held-out mixture, the unit over which every generative comparison is paired, and an \emph{arm} is one source of latent codes, $z$, evaluated on those cells. Generative models of weights can memorize their training checkpoints, particularly
when trained on few samples \citep{zeng2025generative, 10.1007/978-3-032-37383-0_31}, although implicit
regularization can delay this \citep{bonnaire2025why}. Since each of our generators
sees at most 85 mixtures, we test for memorization explicitly.

\paragraph{Continuous Ranked Probability Score.}Mean-squared and mean-absolute error are not proper scoring rules for this question: both are optimized by a point mass, so a generator that has memorized the conditional mean outscores the correct conditional. We therefore report the Continuous Ranked Probability Score (CRPS),
\begin{equation}
  \mathrm{CRPS}(F, y) \;=\; \mathbb{E}\,|X - y| \;-\; \tfrac{1}{2}\,\mathbb{E}\,|X - X'|,
  \qquad X, X' \stackrel{\text{iid}}{\sim} F,
\label{eq:crps}
\end{equation}
whose second term is the diversity credit the point statistics lack \citep{Gneiting2007StrictlyPS}; Appendix~\ref{app:proper} gives the screen that admitted it and the measurement that rejects the alternatives. CRPS needs only one observation per condition, and at $k=50$ our flow is light enough that dense sampling is affordable. We evaluate \eqref{eq:crps} in code space per component, and in function space, where the generated cohort at a held-out mixture is a predictive distribution over each domain's perplexity and the real model is a single observation. We report the skill score
$1-\mathrm{CRPS}_{\text{model}}/\mathrm{CRPS}_{\text{marginal}}$ against the zoo's own marginal, bypassing $\pi$ entirely. As such, a CRPS skill of zero indicates that conditioning is useless. Alongside CRPS skill, we also report round-trip errors of compression and decompression through $z$ (see Appendices~\ref{app:proper} and~\ref{app:blocks}).

\paragraph{Functional round trip.}
Latent-code and weight space metrics do not directly correspond to functional capacity; therefore, we sample a latent at the requested mixture, decode it to $\mathbb{R}^D$, load it into GPT-2 and measure the $\Delta \text{PPL}\%$ on five held-out domain slices (see Appendix~\ref{app:functional}). These held out splits source data in equal proportion to avoid confounding scores. Errors are
$\%\Delta\mathrm{PPL}$: per domain,
$100\cdot|\mathrm{ppl}_{\text{gen}}/\mathrm{ppl}_{\text{real}}-1|$, formed per domain and then averaged. 

\paragraph{Retrieval and Linear Baselines}
One stress test of the generator is to demonstrate the capability to zero shot generalize to held out $\pi_i$ mixtures. The bar is $1$-NN \emph{retrieval}, which returns the real Zoo member whose mixture is nearest the request by $L_1$. A generator that has memorized its training rows and returns the member nearest
the request in $L_1$ reduces exactly to this baseline, so beating retrieval requires
more than copying. For a comparison between such a generator and retrieval system to be level, the retrieval pool must also be denied access to the neighborhood of the held out $\pi_i$. 

The cheapest competing account is that $\theta$ is simply a function of $\pi$. We test least-squares surfaces on $\pi$ and a weight \emph{soup}~\citep{modelsoup}, written in \eqref{eq:soup}, which fits nothing at all, its coefficients being the requested mixture itself, so it is simply task arithmetic over the five vertex means \citep{ilharco2023editing}. Three further measurements corroborate the retrieval comparison with independent failure modes, elaborated on in Appendices~\ref{app:density} and~\ref{app:memorisation}  ~\citep{pmlr-v108-meehan20a}.

\section{Experiments}

\subsection{Zoo Analysis}
\label{exp:zoo}

\paragraph{Cosine similarity.}
Raw pairwise cosine similarities are uniformly high; removing the drift inherited from
the shared trunk separates them, leaving branch deltas strongly aligned within a mixture
and near-orthogonal across mixtures at every scale (Table~\ref{tab:pop-cos}).

\begin{table}[t]
\centering
\caption{Branch delta cosine similarity by pool, raw and after removing the drift inherited
from the shared trunk \eqref{eq:pop-gram}. The pools are disjoint. The
random-direction nulls are $\leq\mathcal{O}(1\text{E-4})$ across scales.}
\label{tab:pop-cos}
\small\setlength{\tabcolsep}{4pt}
\begin{tabular}{lrrrrrrr}
\toprule
& & \multicolumn{2}{c}{Mini (51.5M)} & \multicolumn{2}{c}{Small (124.4M)} & \multicolumn{2}{c}{Medium (354.8M)} \\
\cmidrule(lr){3-4}\cmidrule(lr){5-6}\cmidrule(lr){7-8}
pool & pairs & raw & drift rem. & raw & drift rem. & raw & drift rem. \\
\midrule
within-anchor (same $\pi$)        & 30   & $0.9153$ & $0.8920$  & $0.9024$ & $0.8717$  & $0.8505$ & $0.8000$ \\
between-anchor (different $\pi$)  & 160  & $0.2564$ & $-0.0697$ & $0.2743$ & $-0.0468$ & $0.2423$ & $-0.0160$ \\
singleton--singleton              & 3160 & $0.6047$ & $0.1721$           & $0.6172$ & $0.1193$           & $0.6060$ & $0.0844$ \\
anchor--singleton                 & 1600 & $0.3534$ & $-0.1381$          & $0.3568$ & $-0.1276$          & $0.3132$ & $-0.1167$ \\
all pairs                         & 4950 & $0.5141$ & $0.0684$           & $0.5236$ & $0.0387$           & $0.5011$ & $0.0205$ \\
\bottomrule
\end{tabular}
\end{table}

\paragraph{Spectra.}
Effective rank at $k=99$ grows faster in the transformer blocks than in the embeddings (Table~\ref{tab:pop-spectrum}), so variance not captured by $\pi\in\Delta^4$ grows with model size.

\paragraph{Variance Explained by $\pi$.}
Model-level $\eta^2 \geq 0.85$ at every scale (Table~\ref{tab:pop-anova}), but the
twenty anchors it is measured on sit at the vertices of $\Delta^4$, bounding the
influence of $\pi$ from above, not measuring it on the interior where a generator operates.

\begin{table}[t]
\centering
\caption{One-way between/within decomposition over the twenty anchors
($d=5$ mixtures $\times$ $s=4$ replicate branches, $\mathrm{df}=4/15$), by coordinate
region; $\eta^2$ and $\eta^2_{\text{corr}}$ are \eqref{eq:pop-anova} and
\eqref{eq:pop-anova-corr}. ``of $D$'' is the region's share of the parameter vector,
which shifts from embeddings to blocks with scale.}
\label{tab:pop-anova}
\small\setlength{\tabcolsep}{4pt}
\begin{tabular}{lrrrrrrrrr}
\toprule
& \multicolumn{3}{c}{Mini} & \multicolumn{3}{c}{Small} & \multicolumn{3}{c}{Medium} \\
\cmidrule(lr){2-4}\cmidrule(lr){5-7}\cmidrule(lr){8-10}
region & $\eta^2$ & $\eta^2_{\text{corr}}$ & of $D$
       & $\eta^2$ & $\eta^2_{\text{corr}}$ & of $D$
       & $\eta^2$ & $\eta^2_{\text{corr}}$ & of $D$ \\
\midrule
whole model & $0.9416$ & $0.9224$ & $100\%$  & $0.9220$ & $0.8965$ & $100\%$  & $0.8661$ & $0.8230$ & $100\%$ \\
embeddings  & $0.9455$ & $0.9276$ & $51.0\%$ & $0.9451$ & $0.9271$ & $31.7\%$ & $0.9377$ & $0.9172$ & $14.8\%$ \\
blocks only & $0.9112$ & $0.8822$ & $49.0\%$ & $0.8754$ & $0.8353$ & $68.3\%$ & $0.8186$ & $0.7610$ & $85.2\%$ \\
\bottomrule
\end{tabular}
\end{table}

The LOMO fits of \eqref{eq:pop-loo} measure the interior instead
(Table~\ref{tab:pop-loo}). A quadratic surface on $\pi$ is the strongest predictor at every
scale, capturing roughly 15 to 35\% of interior variance. Only designs carrying $H(\pi)$
predict the vertex folds, possibly because entropy is zero exactly at a one-hot mixture. This range sits
far below the $R^2 = 0.948$ the same quadratic basis reaches on c0--c4
(Appendix~\ref{app:blocks}). 

This gap is also a result, it indicates that the whole-delta fit is dragged down by components that carry little variance and are largely member-private, meaning that the predictable portions are the high-variance ones.
\vspace{-8pt}
\paragraph{Functional diversity.}
\label{exp:zoo_fn}


\begin{wraptable}{r}{0.5\textwidth}
\vspace{-4pt}
\centering\footnotesize
\setlength{\tabcolsep}{3pt}
\caption{Pairwise next-token disagreement: fraction of $327{,}680$ held-out positions at
which two members predict a different argmax token, over the same four disjoint pools as
Table~\ref{tab:pop-cos}. Bold marks the extreme pools of each column.}
\label{tab:pop-fn}
\begin{tabular}{lrrr}
\toprule
pool & Mini & Small & Medium \\
\midrule
within-anchor (same $\pi$) & $0.166$ & $0.179$ & $0.189$ \\
between-anchor (different $\pi$)            & $0.411$ & $0.396$ & $0.373$ \\
singleton--singleton                        & $0.195$ & $0.183$ & $0.184$ \\
anchor--singleton                           & $0.345$ & $0.328$ & $0.312$ \\
all pairs                                   & $0.250$ & $0.236$ & $0.232$ \\
\midrule
between/within anchor            & $2.48$ & $2.21$ & $1.97$ \\
\bottomrule
\end{tabular}
\end{wraptable}

Weight-space distance is not functional difference, so we also record the argmax
decoded token at $327{,}680$ held-out next-token positions per member, a comparison used
to separate genuine ensembles from single-trajectory oversampling \citep{izmailov2019averagingweightsleadswider}. Members with different mixtures disagree roughly twice as often as replicates of the same mixture, at every scale. Our flows incur a real $\%\Delta\mathrm{PPL}$ cost but sit nearest the replicate noise
floor of any source tested, including the $1$-NN lookup estimator, with the closed-form
surfaces on $\pi$ and the weight soup of \eqref{eq:soup} progressively worse
(Table~\ref{tab:gen-ladder}).

\begin{table}[t]
\centering
\caption{Every source of weights on the same $85$ held-out mixtures, in
$\%\Delta\mathrm{PPL}$ over the five-domain profile, averaged per mixture. The
\emph{replicate ceiling} (two real models at the same $\pi$) and $k=50$ rank truncation provide an estimated lower bound. Retrieval is blinded to the same neighbourhood
as the generator (see Table~\ref{tab:fn-vertex} for vertex mixtures.).}
\label{tab:gen-ladder}
\small\setlength{\tabcolsep}{4pt}
\begin{tabular}{lrrrrrr}
\toprule
 & \multicolumn{2}{c}{Mini (51M)} & \multicolumn{2}{c}{Small (124M)} & \multicolumn{2}{c}{Medium (355M)} \\
\cmidrule(lr){2-3}\cmidrule(lr){4-5}\cmidrule(lr){6-7}
source of weights & \%$\Delta$PPL & $\times$ ceil & \%$\Delta$PPL & $\times$ ceil & \%$\Delta$PPL & $\times$ ceil \\
\midrule
replicate ceiling (seed noise) & 1.01 & 1.0 & 1.09 & 1.0 & 0.97 & 1.0 \\
$k=50$ rank truncation floor & 1.80 & 1.8 & 1.93 & 1.8 & 2.46 & 2.5 \\
\midrule
\textbf{flow (ours)} & \textbf{4.05} & \textbf{4.0} & \textbf{4.17} & \textbf{3.8} & \textbf{5.11} & \textbf{5.3} \\
retrieval (1-NN, matched blinding) & 5.75 & 5.7 & 5.79 & 5.3 & 6.01 & 6.2 \\
\midrule
quadratic surface on $\pi$ & 6.54 & 6.5 & 7.27 & 6.7 & 7.88 & 8.1 \\
linear surface on $\pi$ & 9.64 & 9.5 & 11.28 & 10.4 & 12.66 & 13.0 \\
bare trunk (no branching) & 12.72 & 12.6 & 11.48 & 10.5 & 11.42 & 11.7 \\
global mean (ignores $\pi$) & 12.43 & 12.3 & 12.66 & 11.6 & 13.57 & 13.9 \\
soup (Eq.~\eqref{eq:soup}) & 14.87 & 14.7 & 15.45 & 14.2 & 22.29 & 22.9 \\
\midrule
vertices-only flow (Tab.~\ref{tab:fn-vertex}) & 25.09 & 24.8 & 35.18 & 32.3 & 35.77 & 36.7 \\
\bottomrule
\end{tabular}
\end{table}

\subsection{Performance of Generated LLMs}
\label{exp:performance_main}


\paragraph{Downstream tasks.}
\label{exp:performance_downstream}
We score HellaSwag at $2{,}000$ questions on ten held-out mixtures per scale, five
one-hot vertices and five interior, against the real members pretrained at those
mixtures \citep{zellers2019hellaswagmachinereallyfinish}. At these scales and a $20D$
token budget the models sit within a few points of chance and accuracy is quantized at
$1/2000$, so the accuracy column can show parity without being able to distinguish it
from a failure to measure (Table~\ref{tab:gen-bench}, Appendix~\ref{app:res:bench}). We
therefore read the log-probability assigned to the correct continuation, which is
continuous, proper, and paired question by question (Table~\ref{tab:gen-bench-ls}).


\begin{table}[t]
\centering
\caption{Gold-continuation log score on HellaSwag, $2{,}000$ questions, zero shot, ten
held-out mixtures per scale, in nats/token; higher is better. \emph{real} is the members
pretrained at that mixture; \emph{flow} is the leave-one-mixture-out generator.
$\Delta$ columns are paired over the five mixtures of a stratum and give the mean
difference and its $t$ ($\mathrm{df}=4$), with $|t| \geq 2$ the separation threshold used
throughout. \emph{rank $50$} is the same real members' codes truncated to rank $50$, the
reconstruction floor, which involves no generator. The accuracy view of the same runs is Table~\ref{tab:gen-bench}.}
\label{tab:gen-bench-ls}
\small\setlength{\tabcolsep}{4pt}
\begin{tabular}{llrrrr}
\toprule
 & stratum & real & flow & flow $-$ real ($t$) & flow $-$ rank $50$ ($t$) \\
\midrule
Mini (51M) & interior & $-5.198$ & $-5.178$ & $+0.0204$ ($+3.40$) & $+0.0074$ ($+1.62$) \\
 & vertices & $-5.228$ & $-5.140$ & $+0.0877$ ($+0.77$) & $+0.0870$ ($+0.76$) \\
\addlinespace
Small (124M) & interior & $-4.685$ & $-4.648$ & $+0.0360$ ($+4.27$) & $+0.0085$ ($+1.52$) \\
 & vertices & $-4.771$ & $-4.611$ & $+0.1601$ ($+0.97$) & $+0.1597$ ($+0.97$) \\
\addlinespace
Medium (355M) & interior & $-4.148$ & $-4.141$ & $+0.0068$ ($+0.62$) & $-0.0150$ ($-3.12$) \\
 & vertices & $-4.273$ & $-4.101$ & $+0.1720$ ($+0.86$) & $+0.1720$ ($+0.86$) \\
\bottomrule
\end{tabular}
\end{table}
\vspace{-12pt}

\subsection{Diversity of Generated LLMs}
\label{exp:gen_diversity}



\begin{wraptable}{r}{0.5\textwidth}
\vspace{-12pt}
\centering\footnotesize
\setlength{\tabcolsep}{3pt}
\caption{Next-token disagreement of generated models on \emph{interior} pools (vertices are never pooled with the interior). Flow samples disagree with their target more than the rank-50 floor but less than 1-NN retrieval and other $\pi$, and less than original zoo members (Table~\ref{tab:pop-fn}). Vertex results are in Table~\ref{tab:fn-vertex}.}
\label{tab:fn-interior}
\begin{tabular}{@{}lrrr@{}}
\toprule
Pool & Mini & Small & Medium \\
\midrule
Flow vs.\ its target        & 0.1322 & 0.1331 & 0.1452 \\
1-NN retrieval vs.\ target  & 0.1680 & 0.1596 & 0.1651 \\
Flow vs.\ other $\pi$       & 0.2056 & 0.1954 & 0.1967 \\
Rank-50 floor               & 0.0931 & 0.0891 & 0.0935 \\
Flow cohort spread          & 0.0825 & 0.1018 & 0.1257 \\
\bottomrule
\end{tabular}
\vspace{-10pt}
\end{wraptable}
With $80$ of $85$ mixtures observed once, $\pi$ is nearly a unique key, so a generator
could satisfy every conditional statistic by simply storing one point per mixture. Since
no dispersion statistic can detect this (Appendix~\ref{app:proper}), we instead use a
structural control: a copier \emph{is} $1$-NN retrieval, and under matched blinding it
can at best tie retrieval, never beat it.

Conditioning on $\pi$ recovers about three fifths of the available skill over the Zoo's
$\pi$-blind marginal, flat across the ladder (Table~\ref{tab:gen-crps}), and the flow's
exact likelihood correctly identifies the requested mixture among all $85$ candidates for
$94$--$97\%$ of held-out members (Appendix~\ref{app:density}). Generated models therefore
behave like models pretrained on the requested mixture, but respond to it less strongly
than real models do: their perplexity slopes against the requested domain weight are
only 42--54\% of the real slopes (Table~\ref{tab:gen-dose}), and their codes are neither
drawn from the same distribution as real codes nor spread with the correct conditional
width.

\begin{table}[t]
\centering
\caption{All $85$ held-out mixtures, one LOMO generator per cell, with rows indicating $n$ samples/cell. \emph{CRPS} \eqref{eq:crps}
scores the whole cohort against the one real model at that mixture, which a round-trip
error cannot do without being minimized by a point mass (Appendix~\ref{app:proper}). Absolute
round-trip errors are in Table~\ref{tab:gen-ladder}.}
\label{tab:gen-crps}
\small\setlength{\tabcolsep}{4pt}
\begin{tabular}{lrrrrrrrrr}
\toprule
 & \multicolumn{3}{c}{CRPS, \%$\Delta$PPL} & \multicolumn{3}{c}{retrieval $-$ flow, \%$\Delta$PPL} & \multicolumn{3}{c}{marginal $-$ flow, CRPS} \\
\cmidrule(lr){2-4}\cmidrule(lr){5-7}\cmidrule(lr){8-10}
 & flow & marginal & \textbf{skill}($\uparrow$) & gap & $t$ & won & gap & $t$ & won \\
\midrule
Mini (51M, n=32) & 3.10 & 8.34 & \textbf{+0.628} & $+1.70$ & $+4.63$ & 61/85 & $+5.23$ & $+10.27$ & 84/85 \\
Small (124M, n=32) & 3.26 & 8.38 & \textbf{+0.611} & $+1.62$ & $+5.38$ & 61/85 & $+5.12$ & $+11.02$ & 83/85 \\
Medium (355M, n=16) & 3.51 & 8.83 & \textbf{+0.603} & $+0.89$ & $+3.13$ & 43/85 & $+5.32$ & $+12.44$ & 82/85 \\
\bottomrule
\end{tabular}
\end{table}


\subsection{Unseen Mixtures and Reduced Integration Steps}
\label{exp:gen_generalization}

\begin{figure}[tbp]
\centering
\includegraphics[width=\textwidth]{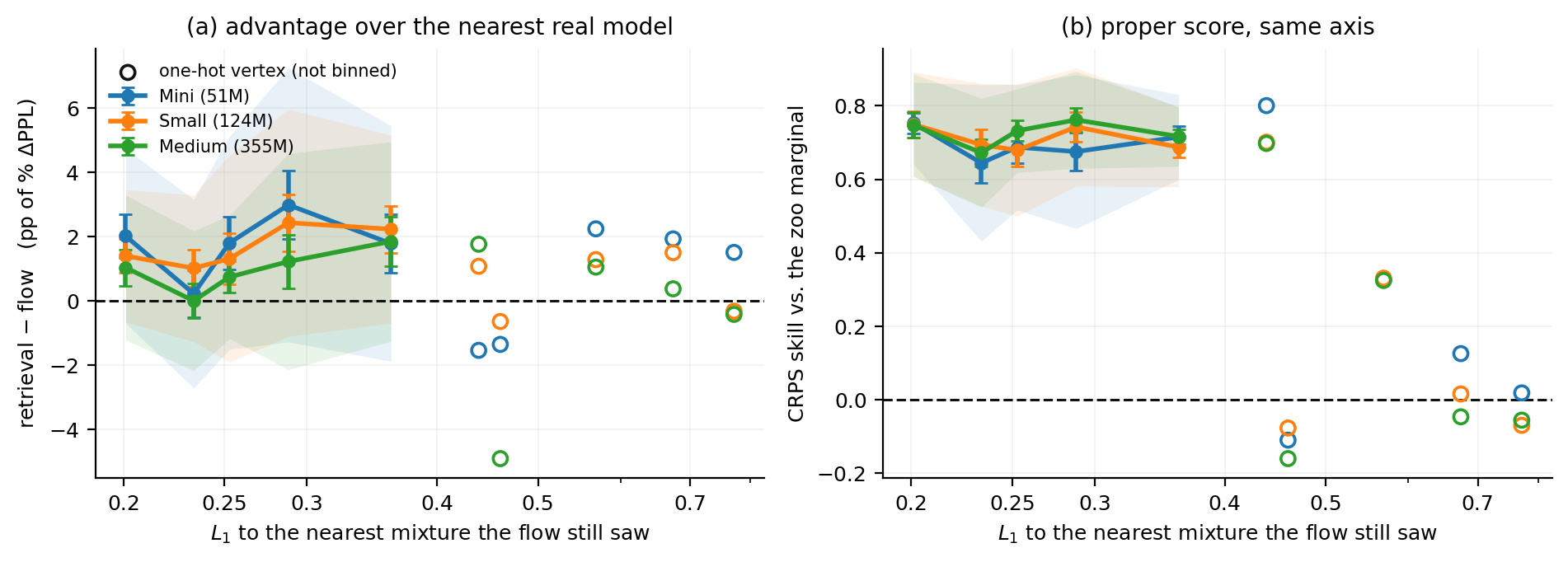}
\caption{Advantage over blinded retrieval (a) and the proper score (b) against $L_1$
distance to the nearest mixture seen by the generator. 
Filled markers are equal-count bins over the $80$ \emph{interior} cells with binned 
standard errors. The five vertices are open squares and are excluded from binning.}
\label{fig:generalization}
\end{figure}


Over the $80$ interior mixtures the margin over blinded retrieval is positive at every
scale and attenuates from Mini to Medium (Table~\ref{tab:gen-strata}).
Figure~\ref{fig:generalization} shows both positive advantage across
distance bins at all three scales and a steady CRPS skill of 
$\sim0.70$. At the five one-hot vertices both quantities drop, with median skill below $0.15$. 

\vspace{-10pt} 
\begin{table}[h]
\centering
\small
\caption{An ablation on the number of Euler integration steps taken in generation of LOMO flows across all scales. Here round-trip error indicates the backward integration, followed by the reverse integration. Mean pairwise distances are reported as a dispersion reporting tool, additional metrics can be found in Table~\ref{tab:n-step-abl-app}.}
\label{tab:step-ablation}
\begin{tabular}{r rrr rrr rrr}\toprule
& \multicolumn{3}{c}{round-trip rel.\ $L_2$} & \multicolumn{3}{c}{rel.\ $L_2$ vs.\ 200-step} & \multicolumn{3}{c}{mean pairwise dist.} \\
\cmidrule(lr){2-4}\cmidrule(lr){5-7}\cmidrule(lr){8-10}
steps & Mini & Small & Med. & Mini & Small & Med. & Mini & Small & Med. \\
\midrule
1 & 1.0077 & 0.7129 & 0.5826 & 0.4158 & 0.3786 & 0.3735 & 2.306 & 3.209 & 4.558 \\
5 & 0.2413 & 0.1552 & 0.1208 & 0.0920 & 0.0715 & 0.0534 & 1.725 & 2.732 & 4.087 \\
50 & 0.0281 & 0.0166 & 0.0122 & 0.0062 & 0.0050 & 0.0037 & 1.922 & 2.889 & 4.139 \\
200 & 0.0070 & 0.0042 & 0.0030 & 0.0000 & 0.0000 & 0.0000 & 1.942 & 2.904 & 4.145 \\
\bottomrule\end{tabular}\end{table}
We find that our flow generator, is capable of quality generation in latent space in very few steps, despite the complexity of the structure. All other results hold the number of Euler integration steps to $50$ unless explicitly stated otherwise.

\section{Conclusion}

We demonstrate that, in a controlled environment, the variance in pretrained language model weight spaces is structured by intentional variation in their pretraining data mixtures, and that this structure is resilient enough to permit our flow matching generative model of weights to generalize inside the simplex of data mixtures. We identify the performance boundaries of these generators, namely the vertices of that simplex, the noise floor of replication, and the baseline performance changes induced by a compression and decompression cycle. We find that generated models perform similarly to zoo members by latent and weight space measures, as well as by downstream task metrics, and that their performance persists through dramatic reductions in the number of generator integration steps. Our methods are limited by the scale of samples used, both in magnitude and population, as well as by the fact that the PCA compression basis is transductive, which limits opportunities for testing basis-blind generation. Finally, we contribute a large, diverse, and linearly mode connected model zoo trained under tight controls for application in weight space learning/generation and beyond. Our implementation is available here: \href{https://github.com/ScottBiggs2/A-Night-at-the-LLM-Zoo}{A-Night-at-the-LLM-Zoo}, and the zoo artifacts are publicly available on \href{https://huggingface.co/collections/ScottBiggs2/gpt-2-zoo}{Hugging Face}.

\subsection*{AI use statement}

In this work, authors leveraged AI tools for the software implementation and adaptation of prior published and appropriately cited methods 
and the data analysis and charting software run on the produced artifacts. 
We did not use AI tools in the decision of methodology, supporting experiments, or analysis of results. Other required disclosure fields are not applicable to our work. We choose to disclose that AI tools were used in the formatting of tables and figures, and for drafting some portions of early manuscript versions. AI tools were also used to support the literature review and source discovery processes; however, authors have verified every cited source for correctness. We take responsibility for the final content of this work, including text, claims or artifacts produced with the aid of generative AI, and will respond to queries regarding the provenance of artifacts, data, methods, or portions of the manuscript text itself. 

\subsection*{Reproducibility statement}

To support the core scientific tenet of peer-reproducibility, we provide all software used in the implementation and execution of this work. 
We support this by detailing relevant hyperparameters throughout the Appendix, including Appendix~\ref{app:zoo_model_details} and Appendix~\ref{app:zoo_model_training_details}. Our implementation is available here: \href{https://github.com/ScottBiggs2/A-Night-at-the-LLM-Zoo}{A-Night-at-the-LLM-Zoo}, and the zoo artifacts are publicly available on \href{https://huggingface.co/collections/ScottBiggs2/gpt-2-zoo}{Hugging Face}.

\subsection*{Ethics statement}

Since several large text corpora were used in training the zoo, we want to thank those who produced, compiled, cleaned, and made publicly available this data. We have cited their work appropriately, and have complied with any relevant licensing and usage terms. To the best of our knowledge, these open source datasets do not reveal human users' personal identifiable information or violate copyright restrictions. 

\subsection*{Acknowledgments}

The authors acknowledge the MGHPCC AI Computing Resource (AICR), with support from the Commonwealth of Massachusetts and the AICR participating institutions, for providing AI and high-performance computing resources used in the results reported in this publication.

This work was supported by the Artificial Intelligence Graduate School Program (KAIST); the National Research Foundation of Korea (NRF) grant funded by the Korean government (MSIT) (No.~RS-2023-00256259); and the InnoCORE program of the Ministry of Science and ICT (No.~N10250156). 

\bibliography{references}
\bibliographystyle{preprint}

\newpage
\appendix

\section{Model Zoo Construction}
\label{app:designing_zoo}

\subsection{Model Architecture}
\label{app:zoo_model_details}

All members of all Zoos are GPT-2 \citep{radford2019language}, built from the Hugging
Face \texttt{GPT2LMHeadModel} with no pretrained weights and no download
(Table~\ref{tab:scales}). The Small and Medium configurations are exactly the published
GPT-2 configurations in both width and depth.

The embedding fraction of $D$ falls by $\sim3.4\times$ across scales because the
transformer blocks grow with $n_{\text{embd}}^2$ while the embedding table grows
linearly. One consequence recurs in every region-wise statistic: a whole-model figure is largely a statement about embeddings at Mini and largely a statement about
transformer blocks at Medium. To account for this evolving ratio, we report statistics for both the blocks and embeddings when appropriate. 

\begin{table}[h]
\centering
\caption{Model scales. $D$ counts every trainable scalar, embeddings and biases
included, and every model is trained on $20D$ tokens \citep{3600270.3602446}. Tokens per
block parameter is the effective data budget on the transformer blocks alone; it swings
$1.74\times$ across the ladder at a fixed $20$ tokens per total parameter.}
\label{tab:scales}
\begin{tabular}{lrrr}
\toprule
 & Mini & Small & Medium \\
\midrule
$n_{\text{embd}}$              & $512$ & $768$ & $1024$ \\
$n_{\text{layer}}$             & $8$   & $12$  & $24$ \\
$n_{\text{head}}$              & $8$   & $12$  & $16$ \\
$D$                            & $51{,}475{,}968$ & $124{,}439{,}808$ & $354{,}823{,}168$ \\
embedding share of $D$         & $51.0\%$ & $31.6\%$ & $14.8\%$ \\
tokens per block parameter     & $40.8$ & $29.3$ & $23.5$ \\
\midrule
\multicolumn{4}{l}{\emph{identical at every scale}} \\
\quad vocabulary               & \multicolumn{3}{c}{$50{,}257$} \\
\quad context length $n_{\text{ctx}}$ & \multicolumn{3}{c}{$1024$} \\
\quad $n_{\text{inner}}$       & \multicolumn{3}{c}{$4\,n_{\text{embd}}$} \\
\quad activation               & \multicolumn{3}{c}{GELU} \\
\quad embeddings               & \multicolumn{3}{c}{input/output tied} \\
\quad LayerNorm $\epsilon$     & \multicolumn{3}{c}{$10^{-5}$} \\
\quad initialization           & \multicolumn{3}{c}{$\mathcal{N}(0, 0.02^2)$, residual projections $\times\,1/\sqrt{2L}$} \\
\quad dropout                  & \multicolumn{3}{c}{$0$ (residual, embedding, attention)} \\
\quad attention                & \multicolumn{3}{c}{fused scaled dot-product} \\
\bottomrule
\end{tabular}
\end{table}

\subsection{Training}
\label{app:zoo_model_training_details}

All $N$ zoo members at a given scale share a common initialization
and a single data stream for a fraction $(1-\beta)S$ of the total step budget, $S$. We refer to this phase of training as the
\emph{trunk}. After $(1-\beta)S$ steps, the zoo population diverges by training on their own mixture $\pi_i$ for the remaining $\beta S$
steps - we call this phase the \emph{branch}. The learning-rate schedule is computed over the global
step count spanning both phases, so branches inherit the optimizer state of the terminal trunk step. 
Table~\ref{tab:zoo-hparams} contains optimizer hyperparameters and Table~\ref{tab:trunkbranch} the resulting step splits.

\begin{table}[h]
\centering
\caption{zoo training hyperparameters, identical at all three scales.}
\label{tab:zoo-hparams}
\begin{tabular}{ll}
\toprule
Optimizer & AdamW \citep{loshchilov2018decoupled} \\
$\beta_{\text{Adam}}$ & $(0.9,\ 0.95)$ \\
Weight decay & $0.1$, applied uniformly \\
Gradient clipping & global norm $1.0$ \\
Precision & bf16 mixed \\
\midrule
Peak learning rate & $6\times10^{-4}$ \\
Warmup & linear, first $1\%$ of steps \\
Decay & cosine to $10\%$ of peak \citep{NEURIPS2020_1457c0d6} \\
\midrule
Batch size & $16$ \\
Gradient accumulation & $8$ \\
Tokens per step & $131{,}072$ \\
Token budget & $20D$ \citep{3600270.3602446} \\
\midrule
Trunk seed & $1234$ (fixed across scales) \\
Branch seed & $10{,}000 + i$ \\
\bottomrule
\end{tabular}
\end{table}

\begin{table}[h]
\centering
\caption{Trunk and branch step split. Counts are exact. $\beta = 0.30$ exists only as the
Mini-scale calibration point of Appendix~\ref{app:beta_selection}.}
\label{tab:trunkbranch}
\begin{tabular}{llrrr}
\toprule
Scale & $\beta$ & total & trunk & branch \\
\midrule
Mini   & $0.15$ & $7{,}854$  & $6{,}676$  & $1{,}178$ \\
Mini   & $0.30$ & $7{,}854$  & $5{,}498$  & $2{,}356$ \\
Small  & $0.15$ & $18{,}988$ & $16{,}140$ & $2{,}848$ \\
Medium & $0.15$ & $54{,}142$ & $46{,}021$ & $8{,}121$ \\
\bottomrule
\end{tabular}
\end{table}

\subsection{Pretraining Data and Zoo Composition}
\label{app:hp-data}

A \emph{class} is a pretraining data mixture $\pi \in \Delta^4$, a point on the simplex
over the five text domains of Table~\ref{tab:domains}. Each domain is one streamed dataset. 
A document is assigned to the held-out evaluation split by
hashing its own text rather than by position: the corpora contain exact duplicate
documents at nearby offsets, so a positional split would let a training document's
duplicate leak into the evaluation set.

Other language models of similar sizes frequently train beyond the $20D$ tok/param ratio, for example SmolLM 135M ~\citet{allal2024SmolLM}, train on over $50\times$ our $20D$ mark. The result is better downstream performance by accepting deviation from ~\citet{3600270.3602446}.

\begin{table}[h]
\centering
\caption{Pretraining data domains. $^\dagger$Python-only because
\texttt{codeparrot-clean} is Python-only. $^\ddagger$French only, one
\texttt{fineweb-2} per-language configuration. Evaluation split is SHA-1 of the UTF-8
document bytes, one in $64$.}
\label{tab:domains}
\small
\begin{tabular}{llll}
\toprule
Domain & Dataset & Config & Column \\
\midrule
web           & \texttt{HuggingFaceFW/fineweb-edu} \citep{penedo2024the} & \texttt{sample-10BT} & text \\
code$^\dagger$ & \texttt{codeparrot/codeparrot-clean} \citep{codeparrot_clean} & --- & content \\
math          & \texttt{open-web-math} \citep{paster2024openwebmath} & --- & text \\
books         & \texttt{manu/project\_gutenberg} \citep{manu_project_gutenberg} & --- & text \\
multilingual$^\ddagger$ & \texttt{HuggingFaceFW/fineweb-2} \citep{penedo2025fineweb} & \texttt{fra\_Latn} & text \\
\bottomrule
\end{tabular}
\end{table}

\paragraph{Composition.}
\label{app:hp-zoo}
Each zoo has $N = 100$ members over $85$ distinct mixtures: $20$ \emph{anchors}, the
five one-hot vertices of $\Delta^4$ replicated over four independent branches, and
$80$ \emph{singletons} drawn i.i.d.\ $\mathrm{Dirichlet}(1)$ and rejection-sampled at a
minimum pairwise $L_1$ gap of $0.15$. On the realized drawing, the closest singleton pair
is at $L_1 = 0.173$ and the median nearest-neighbor distance is $0.251$. The anchor
replicates are used to capture the noise floor from repeated training for a given $\pi_i$. Anchor replicates are exempt from the gap criterion by construction.

\begin{figure}[h]
\centering
\includegraphics[width=\textwidth]{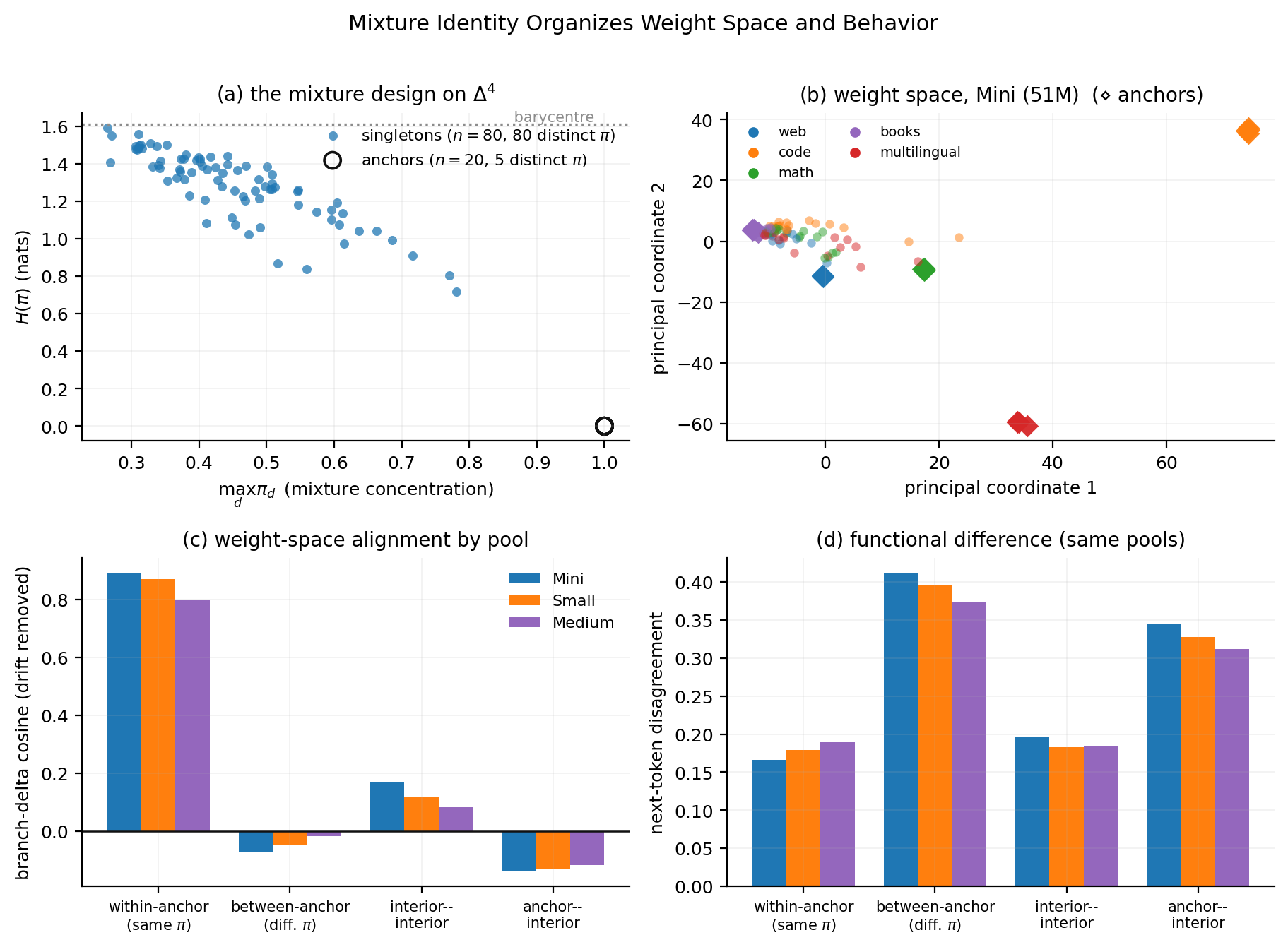}
\caption{\textbf{(a)} The mixture samples shown as entropy, $H(\pi)$ vs the maximum component-wise concentration in $\pi_i$; the design alone, on $\Delta^4$, with no weights involved. \textbf{(b)} The Mini population by top-2 Gram-PCA components, colored by the $argmax \ \pi$ category. \textbf{(c)} The four disjoint pairwise pools of
Table~\ref{tab:pop-cos} in full $\mathbb{R}^D$ weight space. Cosine similarities are read
off the displacement Gram \eqref{eq:pop-gram} and are therefore exact at every coordinate,
without rank truncation. \textbf{(d)} The same four pools in function space, as
next-token disagreement (see Table~\ref{tab:pop-fn}).}
\label{fig:zoo}
\end{figure}

The mixture plan is drawn once and reused, so member $i$ carries an identical
$\pi_i$ at all three scales. The three Zoos are consequently the same experiment at
three sizes rather than three experiments, and a difference between scales is not a
difference in mixtures. As such, the scale tiers do not represent independent replications in the typical sense.

\subsection{Choosing the Branching Fraction}
\label{app:beta_selection}

$\beta$ was calibrated at the Mini scale ($\sim51$M), where a full $N=100$ zoo was
affordable twice-over, against the two candidate values $\beta = 0.15$ and
$\beta = 0.30$. The selection criterion was to find the smallest value $\beta$ that passes the domain-separation gate below, with the reasoning that $\beta$ buys conditioning signal
at the price of compute cost. Both values passed, so $\beta = 0.15$ was chosen in the interest of compute resource efficiency. Only $\beta = 0.15$ was used for the Small and Medium populations, and no $\beta=0.30$ results are included in the main body of this manuscript.

A smaller sweep of $\beta \in \{0.15, 0.30, 0.60\}$ at the Mini scale with $N = 12$ per
setting was run first, to bracket the range before committing a full population; its
weight-space geometry is Table~\ref{tab:beta}.

\label{app:hp-gate}
Mixture identity is checked by scoring every member on held-out text from all five domains, done on $64$ held-out blocks per domain, context length $1024$, and with evaluation RNG seed chosen independently of the training-mixture seeds. The domain separation gate is passed when anchor groups score well in their own domain and the variance between anchor groups exceeds the variance of the groups themselves. This is a loose proxy for the signal-to-noise ratio (SNR) of $\pi$. 
This criterion alone is maximized by memorization or the production of degenerate distributions of weights, and as such it was considered alongside a displacement score w.r.t the trunk artifact (introduced in Sec~\ref{sec:llm_zoo_diversity}), reported against $\beta$ in Table~\ref{tab:beta}. 
We therefore select $\beta$ = 0.15, which is balanced against reasonable trunk displacement and significant compute savings for zoo training and further experiments. Section~\ref{exp:zoo} reports the diversity of the population this produced.

\begin{table}[h]
\centering
\caption{Weight-space geometry against $\beta$, measured at Mini with $N=12$. The
between/within distance ratio is what a PCA basis is built from, and the largest
$\beta$ is the worst of the three on it.}
\label{tab:beta}
\begin{tabular}{lrrr}
\toprule
 & $\beta = 0.15$ & $\beta = 0.30$ & $\beta = 0.60$ \\
\midrule
between/within distance ratio & $\mathbf{4.08}$ & $3.98$ & $2.88$ \\
displacement from trunk & $16.5\%$ & $28.7\%$ & $55.9\%$ \\
\bottomrule
\end{tabular}
\end{table}

\subsection{Common Trunks Canonicalize by Proxy}
\label{app:beta_lmc}

The branching fraction, $\beta$, also decides whether canonicalization is needed. At
$\beta \to 1$ the members are independently trained runs sharing only an architecture,
which is the regime in which weight-space symmetry concerns become unavoidable:
independently trained networks could occupy distinct but functionally equivalent basins
within the loss landscape, related by permutations of hidden units. A coordinate-wise
population statistic would then mix mixture-induced variation with permutation
artifacts, and a basis fitted on the population would inherit that potentially
contaminated mixture. At $\beta \to 0$ the members are one model plus small terminal
perturbations, and the mixture cannot have influenced the weights at all. Neither limit
supports the line of inquiry this paper pursues.

An intermediate $\beta$ is not a compromise between those failure modes, but an
avoidance of both. The mechanism is the shared trunk itself, as all $N$ members enter
the branch phase from identical parameters and only diverge over the final
$\beta S$ steps. Thus the population is generated within a single basin by
construction, not aligned into one afterwards. The alternative is to branch
early or not at all and then re-basin the population before fitting a basis, as prior
weight-space generative work has done \citep{dwflow}; permutation alignment has been
found to buy little for models descended from a shared initialization
\citep{NEURIPS2020_0607f4c7}, and networks that share the first phase of training are
linearly mode connected thereafter \citep{pmlr-v119-frankle20a}; the shared trunk is the
strongest available form of that condition.

\begin{figure}[h]
\centering
\includegraphics[width=\textwidth]{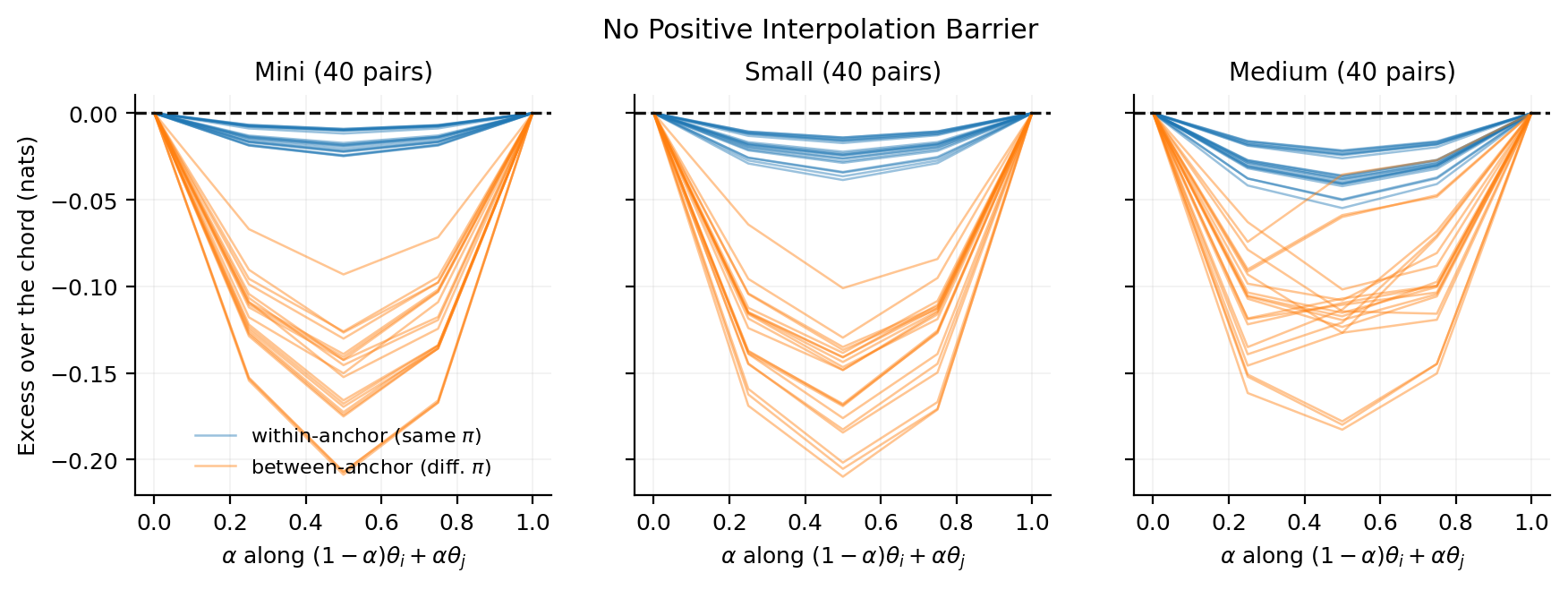}
\caption{Excess of the interpolant's loss over the chord, in nats of mean-over-domain log
perplexity, for all $40$ scored pairs at each scale. Members are interpolated in the full
$\mathbb{R}^D$ weight space, $(1-\alpha)\theta_i + \alpha\theta_j$ on raw weights. Chord endpoints are excluded from reporting.}
\label{fig:barrier}
\end{figure}

Following \citet{pmlr-v119-frankle20a} we score the linear interpolant $\theta(\alpha) = (1-\alpha)\theta_i + \alpha\theta_j$
and report the excess of its loss over the chord, maximized over the \emph{interior} of
the path. The Linear Mode Connectivity (LMC) barrier is negative at every scale and in
both pools: every interpolant lies at least $0.0067$ nats
\emph{below} the chord (Table~\ref{tab:pop-barrier}), which empirically justifies the
decision not to canonicalize the zoo model parameters. Figure~\ref{fig:barrier} draws the
whole path over 40 scored pairs at each scale so total $120$ scored pairs, no interpolant rises above the chord
anywhere.

A single loss basin does \emph{not}, however, mean that the population is interchangeable
within it. Table~\ref{tab:pop-barrier} separates the two: connectivity is scale-stable,
but whether \emph{averaging} two members helps is not, and between different mixtures it
reverses by $355$M, where the midpoint beats both endpoints in only $5$ of $20$ pairs.

\begin{table}[h]
\centering
\caption{Linear-interpolation barrier, in nats of mean-over-domain log perplexity,
maximized over the interior of the path. Negative values mean the interpolated model is
\emph{better} than the chord, so there is no barrier at any scale. The
\emph{midpoint $-$ best endpoint} row is the model-averaging statement and is
independent of the barrier.}
\label{tab:pop-barrier}
\small\setlength{\tabcolsep}{4pt}
\begin{tabular}{lrrrrrr}
\toprule
& \multicolumn{2}{c}{Mini} & \multicolumn{2}{c}{Small} & \multicolumn{2}{c}{Medium} \\
\cmidrule(lr){2-3}\cmidrule(lr){4-5}\cmidrule(lr){6-7}
 & within & between & within & between & within & between \\
\midrule
worst interior barrier            & $-0.0067$ & $-0.0669$ & $-0.0105$ & $-0.0642$ & $-0.0161$ & $-0.0271$ \\
mean dip below chord              & $-0.0165$ & $-0.1568$ & $-0.0239$ & $-0.1587$ & $-0.0353$ & $-0.1251$ \\
midpoint $-$ best endpoint        & $-0.0100$ & $-0.0919$ & $-0.0177$ & $-0.0283$ & $-0.0305$ & $\mathbf{+0.0541}$ \\
midpoint beats both endpoints     & $19/20$ & $18/20$ & $20/20$ & $15/20$ & $20/20$ & $\mathbf{5/20}$ \\
\bottomrule
\end{tabular}
\end{table}

\section{Weight-Space Representation}
\label{app:compression_algo}

\subsection{Dual Gram PCA}

The representation is the dual (Gram) PCA of \citet{dwflow}. Each member is flattened to
$\theta_i \in \mathbb{R}^D$ under a fixed layout, the population is centered as
$\widetilde{\mathbf{W}}$, and the model-space Gram $\mathbf{G} =
\widetilde{\mathbf{W}}^{\top}\widetilde{\mathbf{W}} \in \mathbb{R}^{N \times N}$ is
eigendecomposed into $\mathbf{U}_k$, $\boldsymbol{\Lambda}_k$, giving codes
\begin{equation}
  \mathbf{C} \;=\; \mathbf{U}_k \boldsymbol{\Lambda}_k^{1/2} \in \mathbb{R}^{N \times k}.
\label{eq:dualcodes}
\end{equation}
No $D \times k$ basis is ever materialized, and decoding a code is a weighted sum over
the ensemble members themselves, which is what makes the representation affordable at
$D \sim 10^{8}$. Remaining settings are in Table~\ref{tab:pca-hparams}. Biases are
retained because GPT-2 carries one on every projection, and excluding one-dimensional
parameters would make a generated model inherit some particular member's biases rather
than generate them.

\begin{table}[h]
\centering
\caption{Weight-space representation hyperparameters.}
\label{tab:pca-hparams}
\begin{tabular}{ll}
\toprule
Rank $k$, population statistics & $N - 1 = 99$ \\
Rank $k$, generative experiments & $50$ (Appendix~\ref{app:k_choice}) \\
One-dimensional parameters & retained ($\sim0.1\%$ of $D$) \\
Code normalization & one global scalar to unit training RMS \\
Synthetic weight-space noise & none \\
\bottomrule
\end{tabular}
\end{table}

\subsection{Why $k = 50$ for Generative Experiments}
\label{app:k_choice}

Every generative experiment departs from the $k = N-1$ default and uses $k = 50$,
because at the generator width we commit to it converged far faster. On an $8,000$ epoch sweep at the Mini scale, shown below in Table~\ref{tab:rank}, $k=50$ recovered more structure than $k=99$ in the calibration case (Mini-scale).

\begin{table}[h]
\centering
\caption{Conditional recovery at width $(64,128,64)$, Mini scale, varying only rank and
duration; $100\%$ is a correct conditional. The only $k = 99$ configuration that
converges used a network seven times wider, which Appendix~\ref{app:cfm_details} shows to be
collapse-prone.}
\label{tab:rank}
\begin{tabular}{lrrr}
\toprule
 & $500$ epochs & $8{,}000$ & $20{,}000$ \\
\midrule
$k = 50$ & $27.6\%$ & $90.2\%$ & $\mathbf{101.1\%}$ \\
$k = 99$ & $1.8\%$  & $21.7\%$ & --- \\
\bottomrule
\end{tabular}
\end{table}

The truncation of $k$ comes with a cost on reconstruction quality, measured in Table~\ref{tab:floor} as $\sim2 \  \Delta\mathrm{PPL}\%$ across scales. 

\begin{table}[h]
\centering
\caption{Round-trip error decomposed into what the representation permits and what the
generator adds, $\%\Delta\mathrm{PPL}$ over $85$ held-out cells.}
\label{tab:floor}
\begin{tabular}{lrrr}
\toprule
 & Mini & Small & Medium \\
\midrule
flow round-trip & $4.05$ & $4.17$ & $5.11$ \\
$-$ truncation floor & $1.80$ & $1.93$ & $2.46$ \\
\midrule
$=$ generator's own excess & $\mathbf{2.25}$ & $\mathbf{2.24}$ & $\mathbf{2.65}$ \\
\bottomrule
\end{tabular}
\end{table}

\subsection{The Representation Is Transductive}
\label{app:transductive}

The dual basis is spanned by the zoo and is fitted on all $N$ members, including any
member later used for evaluation. Holding out these members from the basis fitting as well dramatically worsens their displacement relative to included members. The net effect is that the statistical measures become more likely to show the flow generator as memorizing. Developing a compression method that is comparably scalable but free of this weakness remains a challenge for future work.

\subsection{Structure and Noise Inside the Models}
\label{app:blocks}

The mixture nearly determines where a model sits in the PCA latents component coefficient (or 'code') space, but the map
from $\pi$ to that location is strongly nonlinear. On the leading five components,
$\pi$ explains $R^2 = 0.568$ of the variation linearly, $0.786$ once the scalar
$H(\pi)$ is added, and $0.948$ under a full quadratic basis. A closed-form quadratic
ridge regression therefore reaches $R^2 = 0.948$ at no training cost, which is why
its decoding to weights is the baseline every generative result in this paper is
reported against.

That baseline is only well posed because conditional \emph{location} and conditional
\emph{width} occupy different parts of the spectrum (Table~\ref{tab:blocks}). The
leading five components carry $84\%$ of the retained trace and $0.4\%$ of the
within-mixture variance; components c5--c24 carry $80\%$ of the within-mixture
variance and almost none of the trace. Per anchor group the separation is sharper
still, and the components carrying a group's within-variance differ by group: each
anchor's four branches diverge along a nearly private set of directions, which is
what member-specific noise looks like once it has been given its own principal axes.

\begin{table}[h]
\centering
\caption{Code structure at $k = 99$, pooled over the anchor groups. The rank must be
quoted with this table, because the $\lambda$ shares are shares of the \emph{retained}
trace. ``within/ens.'' compares the twenty anchors' within-group variance against all
$100$ members' total, so it exceeds $1$ wherever the anchor groups are more dispersed
than the ensemble. $\eta^2$ is the anchors-only between-group share, bounded in
$[0,1]$ by construction; five one-hot groups saturate a one-way ANOVA on $\Delta^4$,
so read it as a statement that the within-mixture noise is small in that block, not
as evidence that $\pi$ acts linearly --- the $R^2(\pi)$ column measures that.}
\label{tab:blocks}
\begin{tabular}{lrrrrr}
\toprule
Block & $\lambda$ share & within/ens. & of all within & $\eta^2$ & $R^2(\pi)$ \\
\midrule
c0--c3   & $0.803$ & $0.001$ & $0.4\%$  & $0.9998$ & $0.578$ \\
c4       & $0.038$ & $0.004$ & $0.1\%$  & $0.9934$ & $0.368$ \\
c5--c9   & $0.036$ & $0.969$ & $15.0\%$ & $0.4390$ & $0.182$ \\
c10--c24 & $0.036$ & $4.178$ & $65.2\%$ & $0.0283$ & $0.013$ \\
c25+     & $0.087$ & $0.514$ & $19.4\%$ & $0.0144$ & $0.001$ \\
\bottomrule
\end{tabular}
\end{table}

We therefore read conditional location on c0--c4, conditional width on
c5--c24, and score on c0--c24. Components beyond c24 are excluded from scoring:
they behave as member-private noise that no generator can or should reproduce, and
retaining them lets tail mismatch dominate the comparison. Statistics are reported
in blocks rather than per component because individual components are identified
only where the eigenvalue gaps are large --- the third and fourth components form a
near-degenerate pair whose axes rotate between runs.

 \begin{table}[h]
  \centering
  \caption{Member spectrum by coordinate region. Absolute effective rank $k$ is stated because the normalized variant is not
  $k$-invariant. ``var.\ share'' is the region's share of the total centered variance,
  which is the composition confound in one number --- compare it to the region's share of
  $D$ in Table~\ref{tab:pop-anova}.}
  \label{tab:pop-spectrum}
  \begin{tabular}{lrrrrrr}
  \toprule
  & \multicolumn{2}{c}{Mini (8 blk)} & \multicolumn{2}{c}{Small (12 blk)} & \multicolumn{2}{c}{Medium (24 blk)} \\
  \cmidrule(lr){2-3}\cmidrule(lr){4-5}\cmidrule(lr){6-7}
  region, $k=99$ & eff.\ rank & var.\ share & eff.\ rank & var.\ share & eff.\ rank & var.\ share \\
  \midrule
  whole stack      & $4.466$ & $100\%$   & $5.637$ & $100\%$   & $8.725$  & $100\%$ \\
  blocks only      & $7.198$ & $13.4\%$  & $8.872$ & $35.8\%$  & $11.346$ & $64.7\%$ \\
  embeddings only  & $\mathbf{4.026}$ & $86.6\%$ & $\mathbf{4.084}$ & $64.2\%$ & $\mathbf{4.961}$ & $35.3\%$ \\
  \midrule
  \multicolumn{7}{l}{\footnotesize same eigenvalues, embeddings only, read at lower $k$:} \\
  \quad $k=49$     & $3.669$ & & $3.709$ & & $4.419$ & \\
  \quad $k=24$     & $3.449$ & & $3.465$ & & $4.085$ & \\
  \bottomrule
  \end{tabular}
  \end{table}

\subsubsection{Task Arithmetic Baseline}

As used in Table~\ref{tab:gen-ladder}, the \emph{model soup} linear baseline described in Sections~\ref{sec:gen_diversity} can be written as, 

\begin{equation}
\theta_{\text{soup}}(\pi) \;=\; \theta_{\text{trunk}} \;+\; \sum_{j=1}^{5} \pi_j\,\bar d_j ,
\qquad
\bar d_j = \tfrac{1}{4}\sum_{r=1}^{4} d_{jr},
\label{eq:soup}
\end{equation}

which fits nothing at all- it is simply task arithmetic over the five vertex means \citep{ilharco2023editing}. Additional measurements are elaborated on in Appendices~\ref{app:density} and~\ref{app:memorisation}  ~\citep{pmlr-v108-meehan20a}.

\subsection{Supplemental LOMO Results}
\label{app:more-lomo}

A flow of this size $85$ samples of $50$-dimensional latents trains quickly, allowing complete control over the generators training data in experiments. The
design is LOMO (see Section~\ref{sec:gen_diversity}). This holdout does not impact the PCA basis, which is fit to all samples.  

\begin{table}[h]
\centering
\caption{LOMO predictive $R^2$ of the branch delta on $\pi$ \eqref{eq:pop-loo}, $85$
folds over $100$ members: \textbf{interior} scores the $80$ singleton folds,
\textbf{vertex} the $5$ anchor folds; higher is better within $[-1,1]$. Embedding and
function-space regions are in Table~\ref{tab:pop-loo-regions}.}
\label{tab:pop-loo}
\small
\begin{tabular}{llrrrrrr}
\toprule
& & \multicolumn{2}{c}{Mini} & \multicolumn{2}{c}{Small} & \multicolumn{2}{c}{Medium} \\
\cmidrule(lr){3-4}\cmidrule(lr){5-6}\cmidrule(lr){7-8}
region & design & interior & vertex & interior & vertex & interior & vertex \\
\midrule
whole model    & $\pi$            & $-0.0874$ & $-0.1919$ & $\phantom{-}0.0539$ & $-0.2537$ & $\phantom{-}0.0297$ & $-0.2815$ \\
               & $\pi, H(\pi)$    & $\phantom{-}0.0560$ & $\phantom{-}0.2209$ & $\phantom{-}0.1576$ & $\phantom{-}0.1874$ & $\phantom{-}0.0979$ & $\phantom{-}0.1002$ \\
               & $\pi$ quadratic  & $\phantom{-}0.3251$ & $\phantom{-}0.1170$ & $\phantom{-}0.3420$ & $-0.0043$ & $\phantom{-}0.2174$ & $-0.2518$ \\
\addlinespace
blocks only    & $\pi$            & $\phantom{-}0.0463$ & $-0.0959$ & $\phantom{-}0.0295$ & $-0.1354$ & $\phantom{-}0.0355$ & $-0.2348$ \\
               & $\pi, H(\pi)$    & $\phantom{-}0.1079$ & $\phantom{-}0.1940$ & $\phantom{-}0.0846$ & $\phantom{-}0.1563$ & $\phantom{-}0.0847$ & $\phantom{-}0.1162$ \\
               & $\pi$ quadratic  & $0.2433$ & $-0.0818$ & $0.1871$ & $-0.2009$ & $0.1488$ & $-0.3777$ \\
\bottomrule
\end{tabular}
\end{table}

\begin{table}[h]
\centering
\caption{The embedding and function-space regions of Table~\ref{tab:pop-loo}, on the
same $85$ folds and the same three designs. The embedding region is where the quadratic
surface does best --- it is the only region reaching $R^2 > 0.40$ anywhere --- and
function space is where $\pi$ alone does worst.}
\label{tab:pop-loo-regions}
\small\setlength{\tabcolsep}{4pt}
\begin{tabular}{llrrrrrr}
\toprule
& & \multicolumn{2}{c}{Mini} & \multicolumn{2}{c}{Small} & \multicolumn{2}{c}{Medium} \\
\cmidrule(lr){3-4}\cmidrule(lr){5-6}\cmidrule(lr){7-8}
region & design & interior & vertex & interior & vertex & interior & vertex \\
\midrule
embeddings     & $\pi$            & $-0.1217$ & $-0.2036$ & $\phantom{-}0.0728$ & $-0.3084$ & $\phantom{-}0.0145$ & $-0.3480$ \\
               & $\pi, H(\pi)$    & $\phantom{-}0.0427$ & $\phantom{-}0.2242$ & $\phantom{-}0.2142$ & $\phantom{-}0.2017$ & $\phantom{-}0.1331$ & $\phantom{-}0.0772$ \\
               & $\pi$ quadratic  & $\phantom{-}0.3461$ & $\phantom{-}0.1413$ & $\phantom{-}0.4619$ & $\phantom{-}0.0867$ & $\phantom{-}0.4009$ & $-0.0721$ \\
\addlinespace
function space & $\pi$            & $\phantom{-}0.0362$ & $-0.4358$ & $\phantom{-}0.0073$ & $-0.4089$ & $\phantom{-}0.0025$ & $-0.4004$ \\
               & $\pi, H(\pi)$    & $\phantom{-}0.0900$ & $\phantom{-}0.1125$ & $\phantom{-}0.0577$ & $\phantom{-}0.0754$ & $\phantom{-}0.0434$ & $\phantom{-}0.0325$ \\
               & $\pi$ quadratic  & $\phantom{-}0.1443$ & $-0.4932$ & $\phantom{-}0.1129$ & $-0.4811$ & $\phantom{-}0.0703$ & $-0.6571$ \\
\bottomrule
\end{tabular}
\end{table}



Because the vertex mixtures (anchors) are four-fold replicated, we can create five extrapolation points by holding out one such vertex from the flow's training. This is because the we not only target the four removed anchors, but also target their mean and spread as a group - this is the fifth point. By systematically repeating this for the vertecies of $\Delta^4$, we can estimate the relative difficulties of each axis.

\begin{table}[h]
\centering
\caption{The difficulty axis. Rows one to three are rank terciles of the $L_1$ distance
from a request to the nearest mixture its generator still saw, cut on Mini and reused
verbatim at every scale so the same mixtures fall in the same row; the boundaries are
$0.240$ and $0.291$. Rows four and five split the same $85$ mixtures by \emph{type}
instead, which the terciles cannot do. ``gap'' is retrieval minus flow, so positive
favours the generator.}
\label{tab:gen-strata}
\small\setlength{\tabcolsep}{4pt}
\begin{tabular}{lrrrrrrrrr}
\toprule
 & \multicolumn{3}{c}{Mini (51M)} & \multicolumn{3}{c}{Small (124M)} & \multicolumn{3}{c}{Medium (355M)} \\
\cmidrule(lr){2-4}\cmidrule(lr){5-7}\cmidrule(lr){8-10}
stratum & flow & retr. & gap & flow & retr. & gap & flow & retr. & gap \\
\midrule
lower tercile ($n=29$) & 2.60 & 3.63 & +1.03 & 2.21 & 3.39 & +1.18 & 3.20 & 3.72 & +0.52 \\
middle tercile ($n=29$) & 2.38 & 3.94 & +1.56 & 2.48 & 3.61 & +1.13 & 3.24 & 3.66 & +0.41 \\
upper tercile ($n=27$) & 7.40 & 9.96 & +2.55 & 8.10 & 10.71 & +2.61 & 9.18 & 10.99 & +1.81 \\
\midrule
\emph{interior} ($n=80$) & 2.66 & 4.43 & +1.77 & 2.57 & 4.26 & +1.68 & 3.39 & 4.37 & +0.98 \\
\emph{one-hot vertices} ($n=5$) & 26.27 & 26.83 & +0.56 & 29.79 & 30.38 & +0.59 & 32.65 & 32.23 & -0.42 \\
\bottomrule
\end{tabular}
\end{table}

Vertex error is roughly ten times the interior's, and the vertices are also the
most distant cells, so they pile into the upper tercile and a subset balanced on distance
to the nearest surviving mixture can still over-represent them badly.

\begin{table}[t]
\centering
\caption{All $85$ held-out mixtures with one LOMO generator per cell, producing $32$
samples per cell at Mini and Small and $16$ at Medium. CRPS \eqref{eq:crps}
scores the whole cohort against the one real model at that mixture (Appendix~\ref{app:proper});
skill is CRPS against the $\pi$-blind marginal. Absolute
round-trip errors are in Table~\ref{tab:gen-ladder}.}
\label{tab:gen-crps}
\small\setlength{\tabcolsep}{4pt}
\begin{tabular}{lrrrrrrrrr}
\toprule
 & \multicolumn{3}{c}{CRPS, \%$\Delta$PPL} & \multicolumn{3}{c}{retrieval $-$ flow, \%$\Delta$PPL} & \multicolumn{3}{c}{marginal $-$ flow, CRPS} \\
\cmidrule(lr){2-4}\cmidrule(lr){5-7}\cmidrule(lr){8-10}
 & flow & marginal & \textbf{skill} & gap & $t$ & won & gap & $t$ & won \\
\midrule
Mini (51M) & 3.10 & 8.34 & \textbf{+0.628} & $+1.70$ & $+4.63$ & 61/85 & $+5.23$ & $+10.27$ & 84/85 \\
Small (124M) & 3.26 & 8.38 & \textbf{+0.611} & $+1.62$ & $+5.38$ & 61/85 & $+5.12$ & $+11.02$ & 83/85 \\
Medium (355M) & 3.51 & 8.83 & \textbf{+0.603} & $+0.89$ & $+3.13$ & 43/85 & $+5.32$ & $+12.44$ & 82/85 \\
\bottomrule
\end{tabular}
\end{table}

All three designs regress the branch delta $\Delta_i = w_i - w_{\mathrm{trunk}}$ on features of $\pi$ with no intercept, since $\sum_d \pi_d = 1$ means $\pi$ already spans the constant: \emph{linear} uses the five mixture weights; \emph{$+H(\pi)$}
appends the single scalar $H(\pi) = -\sum_d \pi_d \log \pi_d$; \emph{quadratic} uses
the fifteen products $\{\pi_j\pi_k\}_{j\le k}$, which contain the linear span, so the
three are nested. 

Each is fitted by ordinary least squares and LOMO scored: every member sharing a $\pi$ leaves together, so an anchor's four replicates cannot predict one another. The baseline is the training-set mean, refit per fold, which is why $R^2$ is bounded above by $1$ but can go negative
when a design predicts a held-out mixture worse than the mean of the ones it saw.

\subsection{The Conditional Generator}
\label{app:cfm_details}

\subsubsection{Architecture and Training}

The generator is a conditional rectified flow \citep{liu2023flow,lipman2023flow}
over the normalized codes of Appendix~\ref{app:compression_algo}. Training pairs a Gaussian
source $z_0$ with a data code $z_1$, takes the linear interpolant $z_t = (1-t)z_0 +
t z_1$ at $t \sim U(0,1)$, and regresses the velocity field $v_\psi(z_t, t, \pi)$ onto
the constant target $u = z_1 - z_0$; sampling integrates $v_\psi$ from $t=0$ to $t=1$ by
explicit Euler. The mixture enters through a two-layer MLP whose output is concatenated
to the network input, the encoder-MLP pattern of FiLM \citep{3504035.3504518}, and a
learned null token with conditioning dropout gives classifier-free guidance
\citep{ho2021classifierfree}. Table~\ref{tab:flow-hparams} gives the hyperparameters. 

Despite the small size of the generator, it remains performant in part due to the high quality compression of weights into $\mathbb{R}^k$ ~\citep{dwflow}. 

\begin{table}[h]
\centering
\caption{Conditional flow hyperparameters. The generator is deliberately small: it is
what makes an $85$-cell leave-one-mixture-out sweep affordable, since each cell needs an
independently trained flow.}
\label{tab:flow-hparams}
\begin{tabular}{ll}
\toprule
Type & rectified flow \citep{liu2023flow,lipman2023flow} \\
Width & $(64,\ 128,\ 64)$ \\
Parameters & $\sim\!51$k \\
Integration & $50$ explicit Euler steps \\
\midrule
Conditioning & $2$-layer SiLU MLP, concatenated \citep{3504035.3504518} \\
Conditioning dropout & $0.15$ \citep{ho2021classifierfree} \\
Guidance scale at evaluation & $1.0$ \\
Source scale $\sigma_0$ & $1.0$ \\
\midrule
Optimizer & Adam \\
Learning rate & $3\times10^{-4}$ \\
Weight decay & $10^{-4}$ \\
Batch size & $64$ \\
Epochs & $20{,}000$, always run in full; Table~\ref{tab:rank} sweeps it \\
Checkpoint selection & best validation loss, $15\%$ of rows held out \\
Seed & $1234$ \\
\midrule
Training time & $<2$ minutes, CPU \\
\bottomrule
\end{tabular}
\end{table}

Since Table~\ref{tab:tuning} shows that sample dispersion and conditional widths are not simply related, we rank solutions by the proper CRPS statistic and use
$|1 - a_\pi|$ alongside a classifier two-sample test as guards.
$a_\pi$ is two-sided, since over-fitting on a real zoo could appear as $a_\pi$ falling \emph{below} $1$ despite the pooled ratio still seeming healthy.

\begin{table}[h]
\centering
\caption{Flow source width, $\sigma_0$, sweep at established hyperparameter values (see Table~\ref{tab:flow-hparams}). $a_\pi$ is the sampled spread at one
mixture over that mixture's own within-group spread ($[0,1]$, higher is not always better). Column 
``pooled'' is the corresponding ratio against the whole training set - as such they cannot be cleanly compared to the LOMO generators.}
\label{tab:tuning}
\begin{tabular}{lrrrr}
\toprule
& \multicolumn{2}{c}{Mini} & \multicolumn{2}{c}{Small} \\
\cmidrule(lr){2-3}\cmidrule(lr){4-5}
$\sigma_0$ & $a_\pi$ & pooled & $a_\pi$ & pooled \\
\midrule
$0.125$ & $0.316$ & $0.950$ & $0.591$ & $0.992$ \\
$0.25$  & $0.521$ & $0.966$ & $0.485$ & $0.952$ \\
$0.5$   & $\mathbf{0.877}$ & $0.983$ & $\mathbf{0.868}$ & $0.999$ \\
$0.75$  & $0.774$ & $0.983$ & $0.908$ & $0.990$ \\
$1.0$ (committed) & $0.707$ & $0.976$ & $0.681$ & $0.966$ \\
$2.0$   & $0.712$ & $0.959$ & $0.743$ & $0.969$ \\
\bottomrule
\end{tabular}
\end{table}

Modulating every block rather than the input alone, in the manner of adaLN-Zero \citep{10377858}, is worse than concatenation at our chosen width and better than it only with a wider residual stream. None of these improve measured capability. The decided configuration was selected on functional CRPS, and $a_\pi$ is reported as a collapse guard only.

\subsubsection{Step Ablations}

\begin{table}[h]\centering
\caption{Small model scale ablation of varied Euler integration steps with 4 leave-one-mixture-out flows (32
samples). \emph{Round-trip error} integrates a real latent backwards ($1\!\to\!0$) then forwards again ($0\!\to\!1$) and measures how far it
lands from its start. Real latents should sit around $1$ RMS, and this value is included to detect collapsed flows. This table expands on Table~\ref{tab:step-ablation} with additional metrics and dispersion checks.}
\label{tab:n-step-abl-app}
\begin{tabular}{rrrrrrr}\toprule
& \multicolumn{2}{c}{round-trip error} & \multicolumn{2}{c}{vs.\ 200-step sample} & \multicolumn{2}{c}{dispersion} \\
\cmidrule(lr){2-3}\cmidrule(lr){4-5}\cmidrule(lr){6-7}
steps & rel.\ $L_2$ & $\times\,n$ & rel.\ $L_2$ & cosine & sample RMS & mean pair dist. \\\midrule
1 & 0.5902 & 0.590 & 0.2772 & 0.9617 & 1.050 & 4.432 \\
2 & 0.3116 & 0.623 & 0.1346 & 0.9911 & 1.062 & 4.046 \\
5 & 0.1272 & 0.636 & 0.0464 & 0.9989 & 1.058 & 3.982 \\
10 & 0.0633 & 0.633 & 0.0217 & 0.9998 & 1.055 & 3.996 \\
25 & 0.0252 & 0.630 & 0.0079 & 1.0000 & 1.054 & 4.011 \\
50 & 0.0126 & 0.629 & 0.0034 & 1.0000 & 1.053 & 4.017 \\
200 & 0.0031 & 0.628 & 0.0000 & 1.0000 & 1.053 & 4.022 \\
\bottomrule\end{tabular}
\end{table}

\begin{table}[h]\centering
\caption{Small: all 85 LOMO cells, 8
samples per cell with identical seeds. $a_\pi$ is the
conditional dispersion ratio, where $1$ is correct and $<1$ is over-tight. We see that generated models survive drastic reductions in Euler integration steps in function space.}
\begin{tabular}{rrrrr}\toprule
steps & $\Delta$PPL \% & CRPS \% & CRPS skill & $a_\pi$ \\\midrule
1 & 4.365 & 3.134 & 0.6259 & 0.839 \\
10 & 4.165 & 3.252 & 0.6119 & 0.673 \\
50 & 4.173 & 3.231 & 0.6143 & 0.668 \\
\bottomrule\end{tabular}\end{table}



\subsection{Admissible Scores}
\label{app:proper}

As discussed throughout  Section~\ref{sec:methods}, a scoring rule is \emph{proper} if its expected value is optimized by reporting the true
conditional, and \emph{strictly} proper if only by reporting it \citep{Gneiting2007StrictlyPS}. The obvious Mean Squared or Mean Absolute Errors (MSE or MAE) fail to rise to this bar, as they can be minimized by point-mass memorization on the mean alone. This is again the motivation for the CRPS score ~\eqref{eq:crps}, which cannot be minimized by a point mass. To support CRPS and CRPS skill, we report round-trip and functional error measures. 

\subsection{Likelihood and Identification}
\label{app:density}

A divergence between the real and predicted code distributions cannot be convincingly estimated from
samples at our $N$: the $k$-nearest-neighbour estimator returns a negative value on these codes, which is impossible for a non-negative quantity. A continuous normalizing flow has an \emph{exact} likelihood instead, calculated by integrating along the path \citep{NEURIPS2018_69386f6b, lipman2023flow}, 
\begin{equation}
\log p_\psi(z_1 \mid \pi) \;=\; \log p_0(z_0) \;-\; \int_0^1
\operatorname{tr}\!\big(\partial v_\psi / \partial z\big)\, dt ,
\label{eq:cnf}
\end{equation}
evaluated with the exact Jacobian trace and cross-checked against the discrete-map
log-determinant. 

From this construction, two clean properties arise: firstly, a direct connection between held-out log-likelihoods and those of the zoo population through cross-entropy, and secondly, it allows direct rankings of candidates by $\log p_\psi(z \mid \pi)$. In both of these settings, $\pi$-blind approaches or ablations provide lower bounds to interpret against.

\begin{table}[h]
\centering
\caption{Code-space density at $k = 50$. The mutual-information bound compares against a
matched twin --- same width, budget and source scale, conditioning ablated --- so the
difference is not a capacity comparison. Identification ranks the true $\pi$ among all
$85$ candidates. The C2ST is quantised at $1/41$ on its held-out split, so the three
scales are identical, at $40/41$.}
\label{tab:density}
\begin{tabular}{lrrr}
\toprule
 & Mini & Small & Medium \\
\midrule
$I(z;\pi)$ lower bound vs.\ matched twin (nats) & $+18.5$ & $+31.9$ & $+4.9$ \\
\midrule
identification, top-$1$ (chance $1.2\%$) & $\mathbf{96.0\%}$ & $\mathbf{94.0\%}$ & $\mathbf{97.0\%}$ \\
identification, top-$5$ & $99.0\%$ & $98.0\%$ & $99.0\%$ \\
median rank of true $\pi$ (of $85$) & $1.0$ & $1.0$ & $1.0$ \\
tie fraction & $0.000$ & $0.000$ & $0.000$ \\
\midrule
C2ST, joint $(z,\pi)$ & $0.976$ & $0.976$ & $0.976$ \\
NN gen$\to$train / train$\to$train & $0.802$ & $0.730$ & $0.935$ \\
\bottomrule
\end{tabular}
\end{table}

Table~\ref{tab:density} uses these facts to show that mixture identification appears scale-stable. Across candidates and scales, these flows identify the correct held out mixture $\sim95\%$ of the time against a chance rate of about $1\%$. However, because ill fitted models may be able to inflate the information bound, we only refer to this result as a sign indicator. C2ST refers to a classifier two-sample test, where $0.50$ represents that the generated and true distributions are truly indistinguishable. The CRPS and CRPS skill measures reported throughout also tell a similar story in the context of the functional similarity results, indicating that our generators are at least learning the conditional mean, but may not be fully capturing the variance. With this said, the prior result in Table~\ref{tab:gen-ladder} showing that our flows outperform even first-nearest neighbor lookups stands. Better capturing this variance remains an open problem.

\subsection{The Functional Round-Trip}
\label{app:functional}

While latent and weight space measures and statistics can be deeply informative, they ultimately cannot truly capture functional performance on downstream application. To show that the generated models survive up projection into $\mathbb{R}^D$, we sample generated latents, decode them, and use the produced weights to populate a GPT-2 model of the appropriate size. We then measure test-set perplexity across five domain samples, as well as downstream performance on HellaSwag ~\citep{zellers2019hellaswagmachinereallyfinish, radford2019language}. The relevant baselines are the replicate noise floor, and the rank-truncation floor (see Table~\ref{tab:gen-ladder}).

\begin{figure}[h]
\centering
\includegraphics[width=\textwidth]{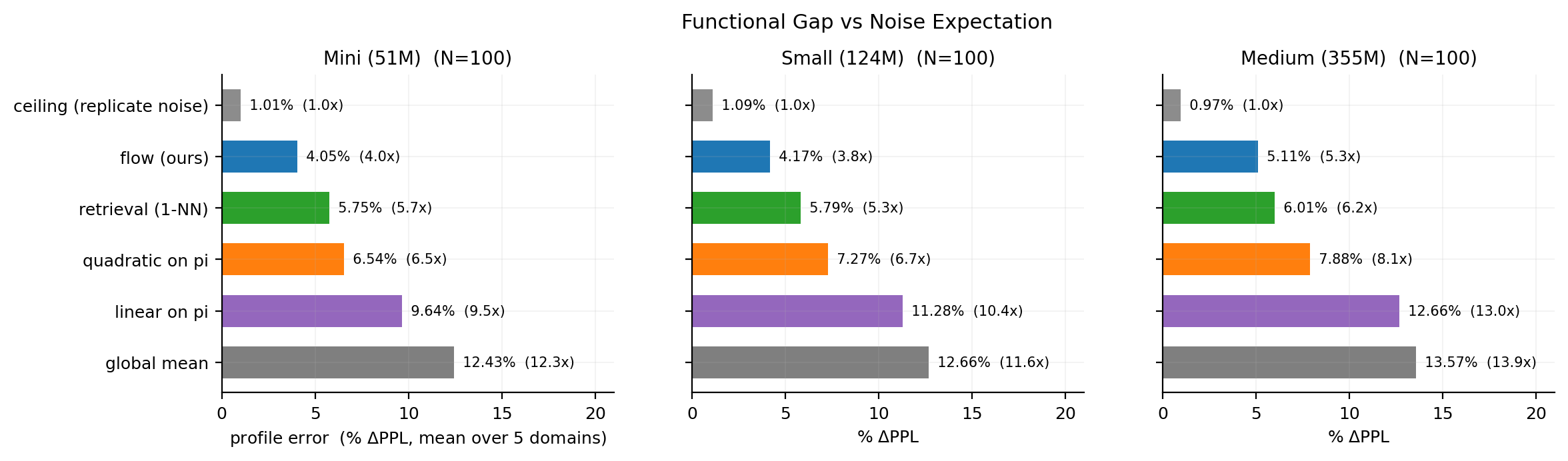}
\caption{Table~\ref{tab:gen-ladder} on one axis per scale, with each bar's multiple of
the replicate ceiling beside it. Every bar is a function-space quantity,
$\%\Delta\mathrm{PPL}$ over the five-domain profile, but the arms reach it differently:
the flow samples a code in $z$ and decodes it to $\mathbb{R}^D$ before evaluation, while
the $\pi$-surfaces, the global mean and retrieval predict the perplexity profile from
$\pi$ directly and never materialize weights. The shape of the ladder holds across a
$6.9\times$ parameter range; what moves is the gap between the flow and retrieval.}
\label{fig:ladder}
\end{figure}

\begin{figure}[h]
\centering
\includegraphics[width=0.72\textwidth]{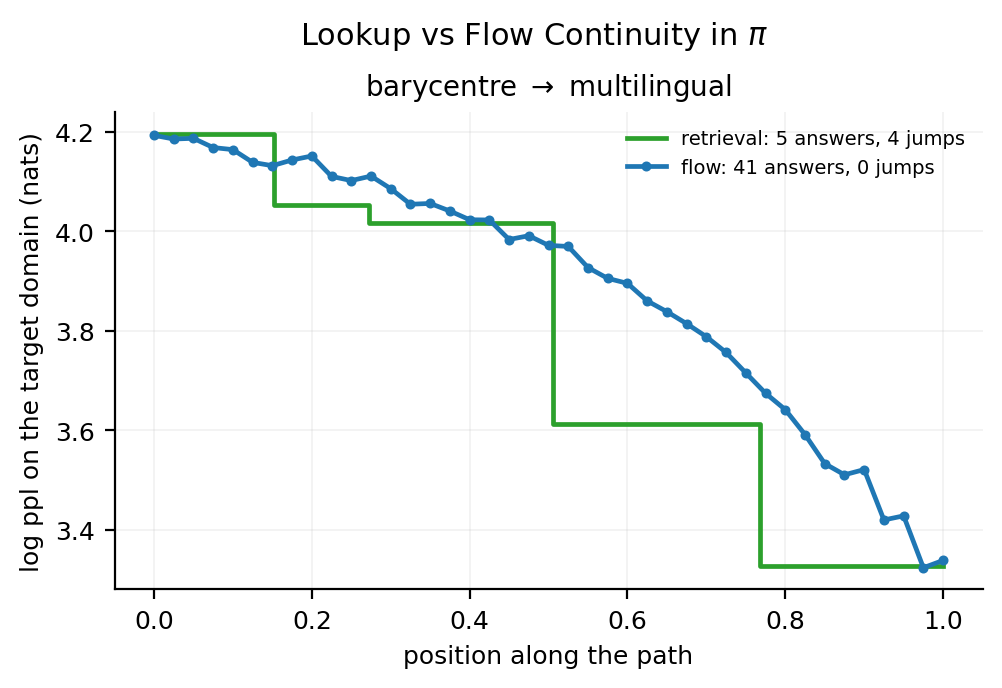}
\caption{One traverse from the barycentre of $\Delta^4$ to the multilingual vertex, $41$
requests at two samples each, with samples at a request averaged before plotting. The
abscissa traverses $\Delta^4$, the ordinate is function space, and each of the flow's
answers is a distinct code in $z$ decoded to $\mathbb{R}^D$ before evaluation.
Retrieval returns five distinct real members over the whole traverse; the generator
returns $41$ distinct models, and both arrive at very nearly the same place.}
\label{fig:continuity}
\end{figure}

The retrieval task along the barycenter-to-vertex traverse is discrete, with discontinuities indicating steps between $\pi_i$ mixtures by 1-NN.  Figure~\ref{fig:continuity} shows this, and the contrast between 1-NN's discrete steps and the flows continuous progress is clear.

\subsection{Dose-Response}
\label{app:dose-response}
\begin{table}[h]
\centering

\caption{Ordinary least-squares slope of log perplexity on domain $d$ against the
requested $\pi_d$, with the Pearson correlation in parentheses, for the real zoo and for
the generated cohort on the same requests. The final row is the mean over domains of the
generated slope as a fraction of the real one.}
\label{tab:gen-dose}
\begin{tabular}{lrrrrrr}
\toprule
 & \multicolumn{2}{c}{Mini (51M)} & \multicolumn{2}{c}{Small (124M)} & \multicolumn{2}{c}{Medium (355M)} \\
\cmidrule(lr){2-3}\cmidrule(lr){4-5}\cmidrule(lr){6-7}
domain & real zoo & flow & real zoo & flow & real zoo & flow \\
\midrule
web & $-0.633$ & $-0.328$ & $-0.754$ & $-0.352$ & $-0.821$ & $-0.325$ \\
code & $-0.635$ & $-0.401$ & $-0.615$ & $-0.349$ & $-0.618$ & $-0.334$ \\
math & $-0.744$ & $-0.458$ & $-0.804$ & $-0.436$ & $-0.861$ & $-0.414$ \\
books & $-0.361$ & $-0.083$ & $-0.431$ & $-0.063$ & $-0.465$ & $-0.058$ \\
multilingual & $-1.056$ & $-0.720$ & $-0.919$ & $-0.535$ & $-0.895$ & $-0.512$ \\
\midrule
mean fraction of real slope & 1.00 & \textbf{0.54} & 1.00 & \textbf{0.46} & 1.00 & \textbf{0.42} \\
\bottomrule
\end{tabular}
\end{table}

\begin{figure}[h]
\centering
\includegraphics[width=\textwidth]{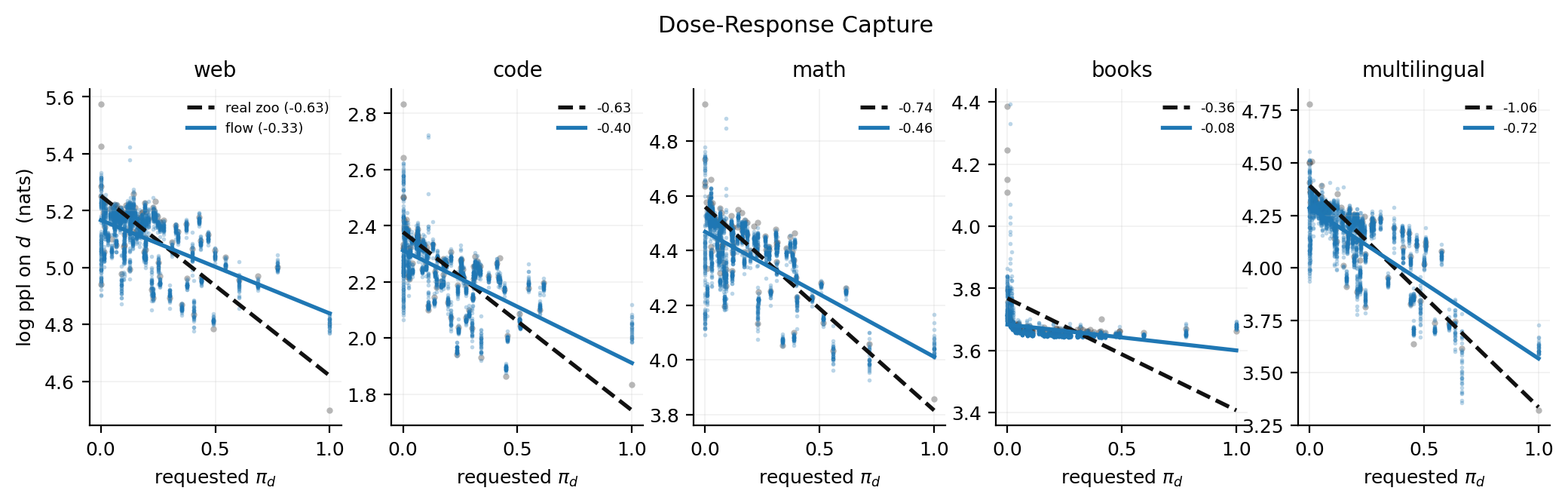}
\caption{The dose--response of Table~\ref{tab:gen-dose} at Mini, measured entirely in
function space: a code is sampled in $z$ at each requested $\pi$ and decoded to
$\mathbb{R}^D$ before any perplexity is taken. Faint points are the
generated cohort ($32$ samples at each of $85$ mixtures) with the real zoo's per-mixture
profiles behind them; the lines carry the slopes of Table~\ref{tab:gen-dose}. The real
slope is fitted per member while the plotted real points are per-mixture means, since an
anchor's four replicates carry leverage their mean does not.}
\label{fig:dose}
\end{figure}

Dose-response curves indicate that across the web, code, math, and multilingual datasets the flows capture a significant proportion of the theoretically possible response to the dose of that dataset. The books category shows low response, which could be due to overlaps with other data classes. Overall, results show that the flows capture a significant proportion of the possible response to $\pi\in\Delta^4$ variations.

\subsection{Next Token Prediction Disagreement}
\label{app:more-next-token}
Here we provide auxiliary results studying the next-token prediction disagreement rates between generated models.

\begin{table}[h]\centering\small
\caption{
Generated model next-token disagreement from \emph{vertex} pools. Vertices and interior are never pooled, organization and comparison follow structurally from $\pi$. Similar results for zoo members
are shown in Table~\ref{tab:fn-interior}. A related Table~\ref{tab:gen-strata} in the main body shows the same experiment run on the
interior of $\Delta^4$. We again find that generated models show higher disagreement rates than noise, but lower
next token disagreement rates than original models share (see Table~\ref{tab:pop-fn}).}
\label{tab:fn-vertex}
\begin{tabular}{lrrr}\toprule
pool & Mini & Small & Medium \\\midrule
flow vs.\ its target & 0.3036 & 0.2991 & 0.2939 \\
1-NN retrieval vs.\ target & 0.3136 & 0.2995 & 0.2887 \\
flow vs.\ other $\pi$ & 0.2744 & 0.2503 & 0.2427 \\
replicate floor (same $\pi$) & 0.1657 & 0.1791 & 0.1892 \\
rank-50 floor & 0.0192 & 0.0133 & 0.0120 \\
flow cohort spread & 0.1302 & 0.1343 & 0.1585 \\
\bottomrule
\end{tabular}
\end{table}

\subsection{Memorization}
\label{app:memorisation}

With $100$ members in a manifold of effective dimension $\sim4.5$, and $80$ of $85$
mixtures observed once, $\pi$ is very nearly a unique key, and a generator could in
principle satisfy every conditional statistic by storing one point per mixture. Four
independent checks bear on it, so memorization would have to defeat all four.

\paragraph{1. The three-sample data-copying test.} Following
\citet{pmlr-v108-meehan20a} we compare nearest-neighbour distances from
\emph{generated} samples to the training set against the same distances from
\emph{real held-out} members, via a Mann--Whitney statistic $Z_U$: a copier gives
$Z_U \ll 0$, an underfit generator $Z_U \gg 0$. The third sample is the set of members the
flow was blinded to, which remain rows of the basis so that their codes are exact;
holding them out of the \emph{basis} instead would inflate their distances and bias the
verdict toward declaring copying (Appendix~\ref{app:transductive}).

\begin{table}[h]
\centering
\caption{Three-sample data-copying test on the leading five components, at four blinded
neighbourhoods and two scales. The mixture plan is identical across scales, so these are
literally the same four mixtures. $Z_U < -3$ is copying, $Z_U > +3$ is underfitting, and
$|Z_U| \le 3$ is generated points sitting where real held-out data sits.}
\label{tab:mcd}
\begin{tabular}{lrrrr}
\toprule
scale & $n_{\text{test}}$ & gen$\to$tr & test$\to$tr & $Z_U$ \\
\midrule
Mini   & $9$ & $0.352$ & $0.435$ & $-2.36$ \\
       & $4$ & $0.511$ & $0.213$ & $+2.59$ \\
       & $9$ & $0.357$ & $0.613$ & $-2.72$ \\
       & $9$ & $0.331$ & $0.292$ & $-1.03$ \\
\addlinespace
Medium & $9$ & $0.671$ & $0.385$ & $+1.99$ \\
       & $4$ & $0.934$ & $0.167$ & $+3.39$ \\
       & $9$ & $0.704$ & $0.572$ & $+1.94$ \\
       & $9$ & $0.668$ & $0.255$ & $+4.03$ \\
\bottomrule
\end{tabular}
\end{table}

No configuration at either scale triggers the copying verdict, and the failure mode that
does appear is underfitting. At Mini the four statistics straddle
zero; at Medium all four are positive and two exceed the underfitting threshold. The sign
change rests on this test alone --- the nearest-neighbour ratio stays below $1$ at Medium
(Table~\ref{tab:density}) and does not corroborate it --- and the test is underpowered:
its third sample is four to nine members and its signs disagree at Mini. So we do not lean
on it. What can be said is that copying is not detected anywhere, and that a memorizing
generator would produce a strongly negative $Z_U$ that never appears.

\paragraph{2--4. Three further checks.} A generator emitting training members
verbatim has zero conditional spread, so its CRPS is dominated by the missing diversity
term and its held-out log-likelihood is unbounded below --- which makes the density of
Appendix~\ref{app:density} the sharpest available collapse control, reported as a control rather
than a headline so that no claim depends on a divergence estimator being correct. The
decisive check is functional. Emitting a copy of the nearest training member \emph{is}
the retrieval baseline, scored on the same cells under the same blinding, so a generator
that copies cannot beat retrieval --- it can only tie it. Any margin over retrieval is
margin copying cannot produce, which is why that comparison is stated per cell and
against extrapolation distance, never as a single average.

\subsection{Downstream Benchmarks}
\label{app:res:bench}

Accuracy on a multiple-choice suite is quantized at $1/n$ and these models sit within a
few points of chance, so Table~\ref{tab:gen-bench} can show parity without being able to
distinguish it from a failure to measure. MMLU is the limiting case and is reported
nowhere: across the $435$ models scored its accuracy takes ten or fewer distinct values,
every one within five questions of $0.242$, while the log-probabilities it assigns
differ on all $2{,}000$ questions --- the argmax over four single-letter continuations
never moves, so the column reports the prior over those four tokens.

\begin{table}[h]
\centering
\caption{HellaSwag accuracy for the runs of Table~\ref{tab:gen-bench-ls}, $2{,}000$
questions, zero shot. \emph{real} is the members pretrained at that mixture, averaged
within a cell before the five mixtures of a stratum are averaged; \emph{flow} is the
leave-one-mixture-out generator at $8$ samples per cell. Accuracy is quantized at
$1/2000 = 0.0005$ and chance is $0.25$.}
\label{tab:gen-bench}
\begin{tabular}{llrrr}
\toprule
 & stratum & real & flow & $\Delta$ \\
\midrule
Mini (51M) & interior & $0.2521$ & $0.2520$ & $-0.0001$ \\
 & vertices & $0.2558$ & $0.2519$ & $-0.0039$ \\
\addlinespace
Small (124M) & interior & $0.2574$ & $0.2580$ & $+0.0006$ \\
 & vertices & $0.2616$ & $0.2590$ & $-0.0026$ \\
\addlinespace
Medium (355M) & interior & $0.2729$ & $0.2715$ & $-0.0014$ \\
 & vertices & $0.2720$ & $0.2680$ & $-0.0040$ \\
\bottomrule
\end{tabular}
\end{table}

The log score of Table~\ref{tab:gen-bench-ls} is the sharper instrument. Scores are near-chance primarily due to the $20D$ training token choice. Models at similar scales, like HuggingFace's SmolLM 135M score $\sim 42\%$, but is trained on almost $250B$ tokens, compared to our $1B$ to $6B$ tokens ~\citep{allal2024SmolLM}.

\end{document}